\documentclass[journal,twoside,web]{ieeecolor}

\usepackage{etoolbox}
\makeatletter
\@ifundefined{color@begingroup}%
{\let\color@begingroup\relax
\let\color@endgroup\relax}{}%
\def\fix@ieeecolor@hbox#1{%
\hbox{\color@begingroup#1\color@endgroup}}
\patchcmd\@makecaption{\hbox}{\fix@ieeecolor@hbox}{}{\FAILED}
\patchcmd\@makecaption{\hbox}{\fix@ieeecolor@hbox}{}{\FAILED}

\usepackage{tmi}
\usepackage{cite}
\usepackage{amsmath,amssymb,amsfonts}
\usepackage{algorithmic}
\usepackage{graphicx}
\usepackage{textcomp}
\usepackage[colorlinks,
            linkcolor=blue,       
            anchorcolor=blue,  
            citecolor=blue,        
            ]{hyperref}
\definecolor{newcolor}{rgb}{.8,.349,.1}
\usepackage{xcolor}
\usepackage{color,colortbl}
\definecolor{mypink}{rgb}{.99,.91,.95}
\definecolor{mygreen}{RGB}{107,147,147}
\definecolor{lgreen}{HTML}{00b8a9}
\definecolor{mygray}{HTML}{eeeeee}
\definecolor{myred}{HTML}{fae4df}
\usepackage{wrapfig}
\usepackage{array}
\usepackage{bbding}
\usepackage{arydshln}
\usepackage{amsmath}
\usepackage{amssymb}
\usepackage{graphicx}
\usepackage{booktabs}
\usepackage{multirow}
\usepackage{hyperref}
\usepackage{makecell}
\usepackage{bbm}
\newcommand{\rr}[1]{\textcolor{red}{#1}}
\newcommand{\bb}[1]{{#1}}
\newcommand{\ib}[1]{{\textit{\textbf{#1}}}}
\newcommand{\s}[1]{}
\definecolor{darkergreen}{RGB}{21, 152, 56}
\newcommand\greenp[1]{\textcolor{darkergreen}{}}
\usepackage[linesnumbered,ruled,vlined]{algorithm2e}

\newcolumntype{I}{!{\vrule width 1.2pt}}
\newlength\savedwidth
\newcommand\whline{\noalign{\global\savedwidth\arrayrulewidth
                            \global\arrayrulewidth 1.5pt}
                   \hline
                   \noalign{\global\arrayrulewidth\savedwidth}}
\newlength\savewidth

\newcommand{\tabincell}[2]{\begin{tabular}{@{}#1@{}}#2\end{tabular}}

\newcommand{\tocite}[1]{{\textcolor{red}{TOCITE}}}

\definecolor{mypink}{rgb}{0,0,0}
\newcommand{\ud}[1]{\textcolor{mypink}{#1}}

\def\BibTeX{{\rm B\kern-.05em{\sc i\kern-.025em b}\kern-.08em
    T\kern-.1667em\lower.7ex\hbox{E}\kern-.125emX}}
\begin{document}
\title{MiM: Mask in Mask Self-Supervised Pre-Training for 3D Medical Image Analysis}
\author{Jiaxin Zhuang, Linshan Wu, Qiong Wang, Peng Fei, Varut Vardhanabhuti, Lin Luo, and Hao Chen,~\IEEEmembership{Senior Member, IEEE}
\thanks{ This work was supported by Hong Kong Innovation and Technology Fund (Project No. ITS/028/21FP, ITCPD/17-9 and MHP/002/22), Shenzhen Science and Technology Innovation Committee Fund (Project No. SGDX20210823103201011), and the Project of Hetao Shenzhen-Hong Kong Science and Technology Innovation Cooperation Zone (HZQB-KCZYB-2020083).\\
    \indent Jiaxin Zhuang and Linshan Wu are with the Departments of Computer Science and Engineering, Hong Kong University of Science and Technology, Hong Kong (email: \{jzhuangad,linshan.wu\}@cse.ust.hk)\\ 
    \indent Qiong Wang is with Shenzhen Institutes of Advanced Technology, Chinese Academy of Sciences, China. (email: wangqiong@siat.ac.cn).\newline
    \indent Peng Fei is with the School of Optical and Electronic Information-Wuhan National Laboratory for Optoelectronics, Huazhong University of Science and Technology, Wuhan, 430074, China (email: feipeng@hust.edu.cn).\newline
    \indent Varut Vardhanabhuti is with the Department of Diagnostic Radiology, The University of Hong Kong, Hong Kong SAR. (email: varv@hku.hk).\newline
    \indent Lin Luo is with the College of Engineering, Peking University, Beijing, China. (email:luol@pku.edu.cn).\newline
    \indent Hao Chen is with the Department of Computer Science and Engineering , Department of Chemical and Biological Engineering, and State Key Laboratory of Molecular Neuroscience, Hong Kong University of Science and Technology, Hong Kong, and HKUST Shenzhen-Hong Kong Collaborative Innovation Research Institute, Futian, Shenzhen, China (corresponding author: jhc@cse.ust.hk). 
}
}

\maketitle

\begin{abstract}
The Vision Transformer (ViT) has demonstrated remarkable performance in Self-Supervised Learning (SSL) for 3D medical image analysis. \ud{Masked} AutoEncoder (MAE) for feature pre-training can further unleash the potential of ViT on various medical vision tasks.
However, due to large spatial sizes with much higher dimensions of 3D medical images, the lack of hierarchical design for MAE may hinder the performance of downstream tasks.
In this paper, we propose a novel \textit{Mask in Mask (MiM)} pre-training framework for 3D medical images, which aims to advance MAE by learning discriminative representation from hierarchical visual tokens across varying scales. We introduce multiple levels of granularity for masked inputs from the volume, which are then reconstructed simultaneously ranging at both fine and coarse levels. Additionally, a cross-level alignment mechanism is applied to adjacent level volumes to enforce anatomical similarity hierarchically. Furthermore, we adopt a hybrid backbone to enhance the hierarchical representation learning efficiently during the pre-training. MiM was pre-trained on a large scale of available 3D volumetric images, \textit{i.e.,} Computed Tomography (CT) images containing various body parts. Extensive experiments on twelve public datasets demonstrate the superiority of MiM over other SSL methods in organ/tumor segmentation and disease classification. We further scale up the MiM to large pre-training datasets with more than 10k volumes, showing that large-scale pre-training can further enhance the performance of downstream tasks. Code and checkpoint will be available once upon acceptance.
\end{abstract}


\vspace{-7pt}
\section{Introduction}\label{sec:intro}
\ud{The advent of deep learning has catalyzed unprecedented advances in medical image analysis, predominantly within the supervised learning paradigm. However, this paradigm's fundamental limitation lies in its reliance on extensively labeled training data, particularly challenging for 3D medical images where annotation demands substantial domain expertise, time, and resources~\cite{tajbakhsh2020embracing,He2024FoundationMF,bassi2024touchstone,jin2023label,yang2023semi,zhuang2023class}. Self-Supervised Learning (SSL) has emerged as a transformative solution, offering a powerful mechanism to learn robust, transferable representations from unlabeled data that significantly enhance downstream task performance~\cite{jiang2023anatomical,he2023geometric,tang2022self,wu2024large,VoCo,ni2024mg,wu2024freetumor,li2024smvlm,ma2024automatic,xu2024multimodal}. This paradigm shift has enabled unprecedented advances in medical image analysis while substantially reducing the annotation burden~\cite{He2024FoundationMF,jin2023label}}.

\ud{In the realm of SSL, Masked Autoencoders (MAE)~\cite{he2022masked} have demonstrated remarkable success in natural image analysis through their innovative approach to reconstructing distorted views via Vision Transformer architectures~\cite{he2022masked,chen2023mixed}. However, extending these methods to 3D medical imaging introduces unique challenges due to the inherent complexity of volumetric data, characterized by high dimensionality and intricate anatomical structures spanning multiple scales. Recent advances have made significant strides in addressing these challenges. MAE3D~\cite{chen2023masked} pioneered the adaptation to 3D medical imaging by reconstructing non-overlapping patches from cropped sub-volumes, establishing a foundation for understanding complex anatomical structures. GL-MAE~\cite{zhuang2023advancing} further advanced this direction by introducing local-global contrastive learning to enhance multi-granular anatomical structure modeling, while SwinUNETR~\cite{tang2022self} incorporated multi-scale feature learning through hybrid transformers~\cite{hatamizadeh2021swin}. Although these approaches demonstrated promising results, their reliance on single-scale cropped volumes inherently constrains their ability to capture comprehensive anatomical relationships. Recent efforts have explored alternative strategies to address these limitations: SwinMM~\cite{wang2023swinmm} leveraged multi-view consistency to enhance volume-wise representations, while Alice~\cite{jiang2023anatomical} utilized pre-trained models~\cite{yan2022sam} to capture rich intra-volume relationships. Despite these advances, three critical challenges persist: the spatial context limitations imposed by volume cropping, which hinder the understanding of complete anatomical structures; the lack of explicit modeling of hierarchical relationships across different anatomical scales; and the substantial computational demands of processing full-resolution 3D volumes. These interconnected challenges underscore the need for a more sophisticated approach that can efficiently capture both fine-grained anatomical details and global contextual information while maintaining practical computational requirements.}

\begin{figure}[ht]
    \begin{center}
    \vspace{-5pt}
	\includegraphics[width=1\linewidth]{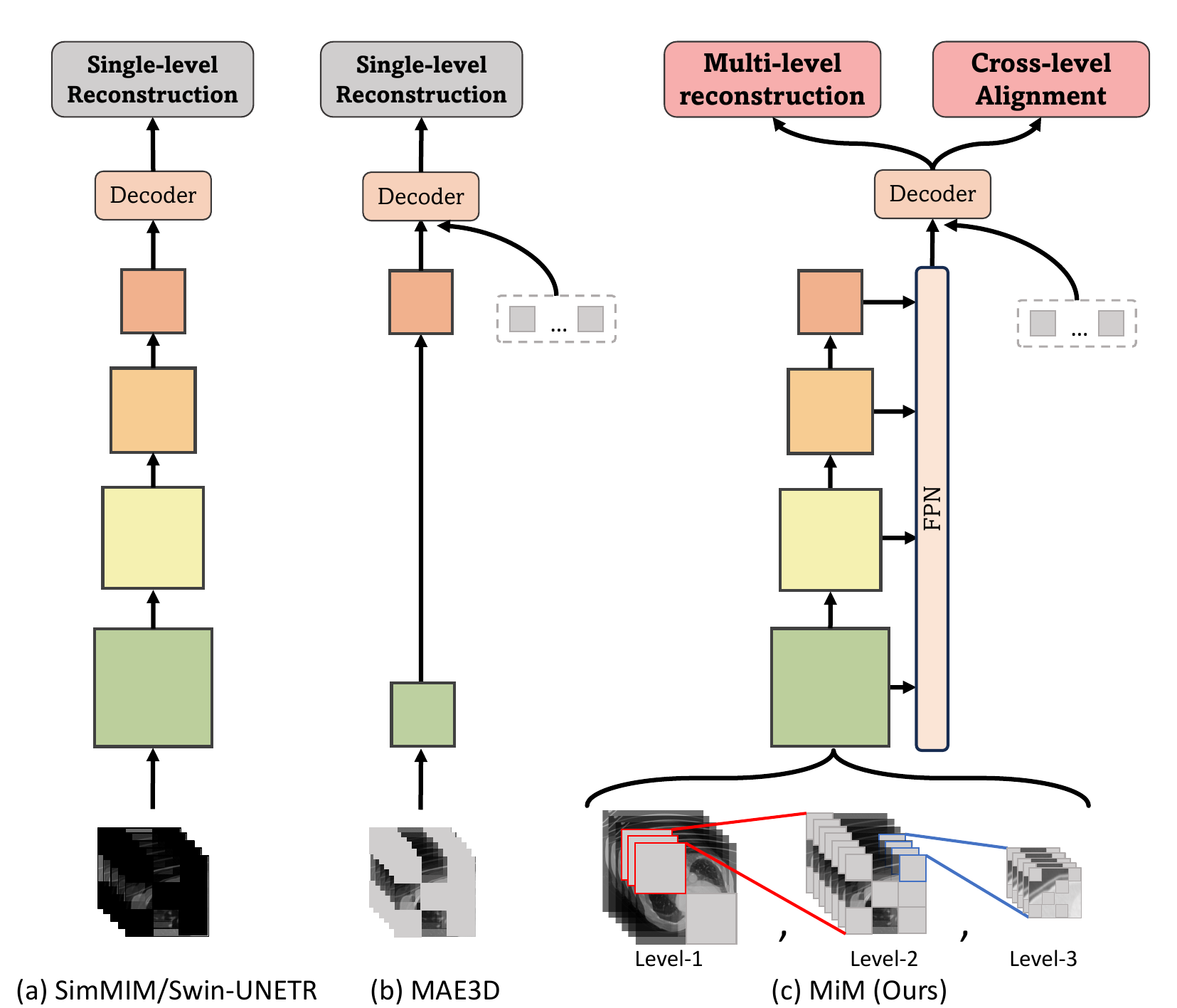}
    \end{center}
    \vspace{-15pt}
    \caption{
        Different SSL for 3D medical image analysis. Current Masked Image Modeling methods for 3D medical images primarily (a) rely on pretext tasks \textit{e.g.,} inpainting, at a \ib{single level}, utilizing hybrid transformers to incorporate all tokens or (b) employ an MAE that reconstructs at a \ib{single level} using \ud{unmasked} tokens. In contrast, (c) we observe that 3D medical images inherently exhibit hierarchical properties. Thus, our Mask in Mask (MiM) framework aims to encode \ib{multi-level} 3D medical image learning across hierarchical visual tokens at various scales through \ib{multi-level reconstruction} and \ib{cross-level alignment} (we set the number of level $L$ to 3 in this figure). Additionally, our framework employs a hybrid transformer while only using \ud{unmasked} tokens.
    }
    \label{fig:highlight}
    \vspace{-10pt}
\end{figure}

\ud{The intrinsic hierarchical nature of medical images, particularly those with expansive spatial dimensions, necessitates a sophisticated multi-level analytical framework for comprehensive clinical interpretation~\cite{chen2022scaling,tang2022self}. As shown in Fig~\ref{fig:highlight}, we introduce MiM, a novel hierarchical framework that fundamentally advances MAE-based representation learning for 3D medical images. Our framework systematically addresses the key limitations of existing approaches through three synergistic components. First, to overcome the limited context of cropped volumes, we propose a multi-level volume generation strategy that processes the larger view of 3D volume at multiple scales simultaneously, enabling our model to capture both fine-grained anatomical details and their broader contextual relationships. Second, to explicitly model hierarchical representations, we design a sophisticated multi-level reconstruction mechanism that operates across different anatomical scales. This mechanism preserves critical anatomical details at varying granularities while enforcing consistency through an advanced cross-level alignment strategy, ensuring coherent interpretation between local structures and their global context. Third, to address the computational challenges of processing larger view of the 3D medical images, we incorporate an efficient hybrid backbone design inspired by MCMAE~\cite{gao2022convmae}. This architecture significantly reduces computational overhead while maintaining the advantages of transformer-based models, making it practical to analyze high-resolution 3D medical images. These innovations work together synergistically, enabling MiM to effectively model the complete anatomical hierarchy while maintaining computational efficiency.}

\ud{The principal contributions of this work are threefold:
\begin{enumerate}
    \item We present MiM, a computationally efficient SSL framework that advances MAE through hierarchical design for 3D medical image pre-training. Our approach effectively manages the complexity of 3D medical data while enabling the simultaneous capture of anatomical features across multiple scales, crucial for accurate medical image analysis.
    \item We introduce a comprehensive methodology for encoding multi-level visual information through two synergistic proxy tasks: multi-level reconstruction and cross-level alignment. This design enables robust local and global representation learning while maintaining anatomical consistency across scales through our novel cross-level alignment mechanism.
    \item Through extensive experimental validation across twelve diverse datasets, utilizing pre-training sets ranging from \ib{1k} to \ib{10k} volumes, we demonstrate state-of-the-art performance and establish a clear correlation between pre-training dataset scale and model effectiveness. Our comprehensive evaluation shows significant improvements across various medical imaging tasks, with our multi-level approach consistently outperforming single-scale alternatives.
\end{enumerate}}

\section{Related works}\label{eq:relatedworks}
\ud{Recent advances in SSL for medical image analysis have evolved through different paradigms, with dense prediction tasks like segmentation being crucial yet challenging for 3D medical images. In this section, we first review SSL methods in medical imaging, focusing on the progression from contrastive-based to generative-based approaches and their effectiveness in such dense prediction tasks. This evolution leads to our discussion of Masked Image Modeling, a powerful generative SSL technique showing promise in 3D medical image analysis. However, current Masked Image Modeling approaches struggle to capture the complex hierarchical nature of 3D medical data, motivating our examination of hierarchical SSL designs. Through this review, we demonstrate how existing methods fall short in effectively modeling multi granularity of features in 3D medical images, thereby establishing the context for our proposed hierarchical MIM framework.}\\
\noindent\textbf{SSL for Medical Image Analysis.}
\ud{SSL methods in medical imaging have evolved through two primary paradigms: contrastive-based and generative-based approaches~\cite{du2024teach,gao2022disco,gui2024survey}. \textit{Contrastive-based} methods focus on aligning representations of positive pairs while separating negative pairs in feature space~\cite{chen2020simple}. While foundational works like SimCLR~\cite{chen2020simple}, MoCov2~\cite{chen2020mocov2}, and MoCov3~\cite{chen2021empirical} demonstrated success in medical classification tasks~\cite{zhou2021preservational,ye2022desd}, their effectiveness in medical imaging has been further enhanced through domain-specific innovations. For instance, ~\cite{taleb20203d} introduced sub-volume based contrastive learning, while ~\cite{xie2020pgl,yan2022sam} advanced multi-granular understanding by contrasting features at both local and global scales. However, these approaches' focus on instance-level alignment has limited their effectiveness in dense prediction tasks like segmentation~\cite{zhou2021preservational}. \textit{Generative-based} approaches address these limitations by explicitly modeling spatial structures and preserving local-global consistency~\cite{zhu2020rubik}. The field has progressed from basic reconstruction tasks using 3D UNet~\cite{ronneberger2015unet,zhou2021models} to sophisticated approaches incorporating modern architectures like 3D Swin Transformers with inpainting~\cite{tang2022self,taleb20203d}. This evolution has culminated in Masked Image Modeling, demonstrating robust performance across various 3D medical imaging applications~\cite{he2022masked,xie2022simmim,chen2023masked,zhuang2023advancing,prabhakar2024vit}.}

\noindent\textbf{Medical Image Analysis with Masked Image Modeling.}
\ud{Masked Image Modeling has emerged as a powerful technique for medical image analysis, building upon the success of generative SSL. Initially proposed by~\cite{he2022masked}, this approach demonstrated that high-ratio masking creates an effective self-supervisory task through raw pixel restoration. The field has advanced through innovations in masking strategies~\cite{wang2023hard} and reconstruction targets~\cite{chen2023mixed}. While early applications showed promise in 2D medical tasks~\cite{zhang2023dive,wang2024fremim}, including disease classification~\cite{xiao2023delving} and segmentation~\cite{wang2024fremim}, the extension to 3D medical imaging introduced new challenges. MAE3D~\cite{chen2023masked} pioneered the adaptation of masked autoencoding to volumetric data, while GL-MAE~\cite{zhuang2023advancing} enhanced anatomical structure understanding by incorporating global-local consistency through contrastive learning. SwinUNETR~\cite{tang2022self} further advanced the field by recognizing the importance of multi-scale representation learning and incorporating hybrid transformers~\cite{hatamizadeh2021swin}. However, these methods~\cite{chen2023masked,zhuang2023advancing,tang2022self,wang2023swinmm}, despite their innovations, process 3D medical images at a single scale with limited receptive fields or rely solely on backbone-level multi-scale processing. As illustrated in Fig.~\ref{fig:highlight}, this approach becomes problematic when handling the significantly larger spatial dimensions and varying anatomical structures present in 3D medical images.}

\noindent\textbf{Hierarchical SSL.}
\ud{The importance of hierarchical structures in SSL has gained recognition in both natural and medical image domains. In natural image processing, approaches like~\cite{chen2023mixed} and~\cite{xie2022simmim} have integrated Masked Image Modeling with hierarchical backbones like Swin Transformer~\cite{liu2021swin}.~\cite{wang2023masked} explored multi-scale feature reconstruction, while~\cite{gao2022convmae} advanced masked patch prediction using unmasked contexts. In medical imaging, Adam~\cite{hosseinzadeh2023towards} introduced multi-granular contrastive learning with ResNet~\cite{he2016deep}. However, these approaches either suffer from the limited dimensionality of 2D images~\cite{hosseinzadeh2023towards,gao2022convmae} or treat hierarchical levels independently~\cite{hosseinzadeh2023towards}, failing to capture the cross-level semantic relationships crucial for 3D medical images.}\\
\ud{Distinguished from these approaches, our MiM framework advances the state-of-the-art in 3D medical image analysis through four key innovations: (1) We radically advance beyond MAE3D~\cite{chen2023masked} by designing a powerful multi-level volume generation strategy that enables simultaneous reconstruction of features across both fine and coarse scales, addressing the limitations of existing single-scale methods with limited receptive fields~\cite{chen2023masked,tang2022self,wang2023swinmm,zhuang2023advancing} (2) Inspired by ~\cite{caron2021emerging,zhuang2023advancing}, we develop a sophisticated cross-level alignment mechanism specifically for 3D medical volumes that enforces anatomical consistency across hierarchical levels, achieving significant improvements in anatomical feature learning. (3) Our innovative 3D hybrid backbone architecture adapted from 2D natural images~\cite{gao2022convmae} achieves superior efficiency in capturing multi-scale features during pre-training while reducing computational demands. (4) MiM demonstrates exceptional scalability by successfully pre-training on massive datasets exceeding 10,000 volumes, surpassing the scale of existing 3D SSL medical imaging methods~\cite{zhou2021models,zhou2021preservational,chen2023masked,zhuang2023advancing,tang2022self,he2023geometric}. The effectiveness of these innovations is validated through extensive experiments across twelve public datasets in organ/tumor segmentation and disease classification tasks, establishing MiM as a transformative advancement in the field.}
\section{Methodology}\label{sec:method}
This section provides an overview of our proposed MiM method. Firstly, the overall framework of the MiM method is introduced in Section~\ref{sec:framework}. Secondly, the process of multi-level reconstruction is presented in Section~\ref{sec:multi_reconstruction}. Then, the cross-level alignment via contrastive learning in our proposed MiM method is described in Section~\ref{sec:cross}. Finally, the backbone of the MiM method in the pre-training period is introduced in Section~\ref{sec:backbone}.
\begin{figure*}[!ht]
	\centering
    \vspace{-10pt}
	\includegraphics[width=1\linewidth]{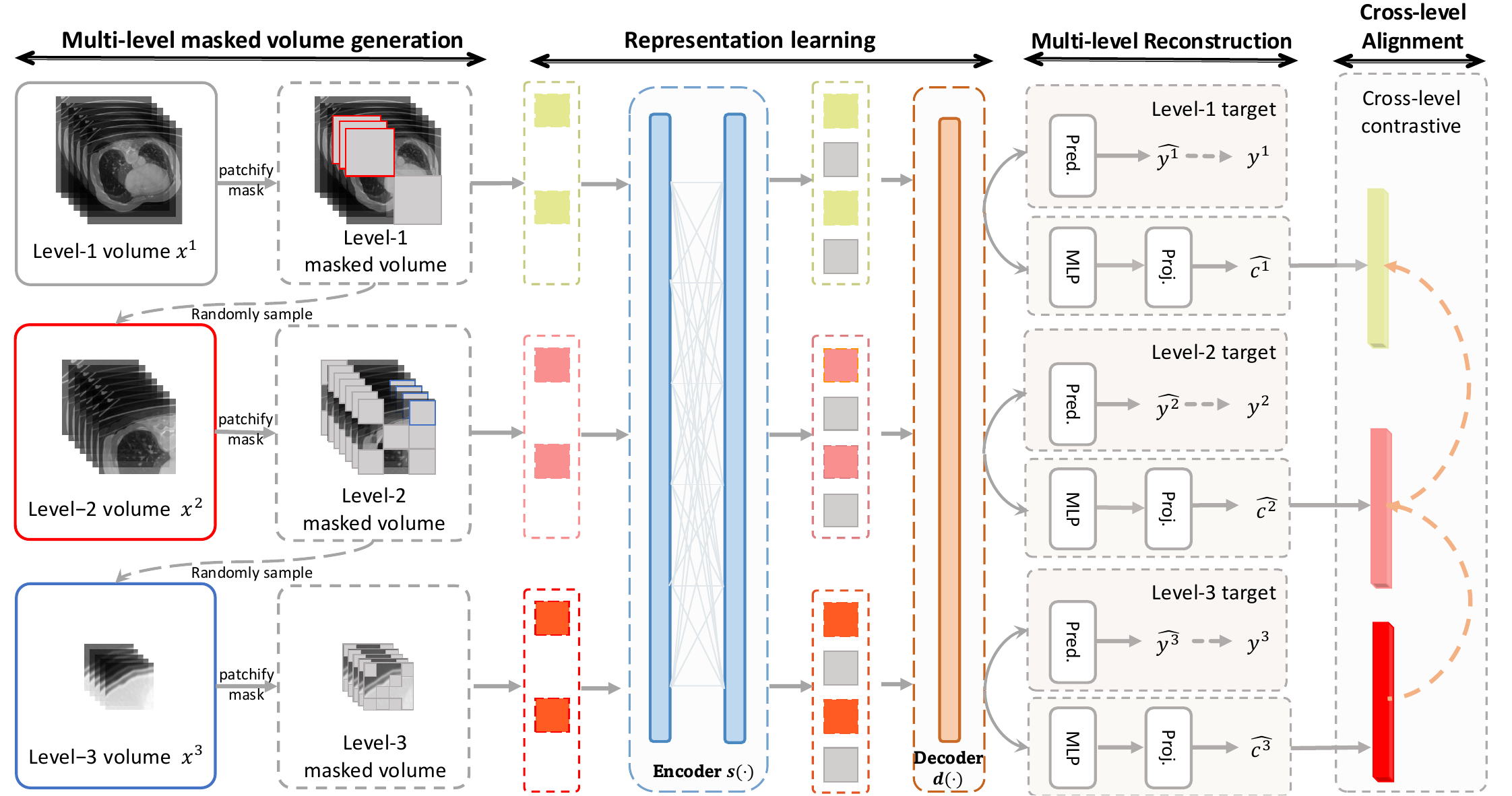}
	\vspace{-20pt}
	\caption{\ud{The overall view of our MiM pre-training framework. The level $L$ is set to 3 for better illustration. We first conduct the process of multi-level masked volume generation. The multi-level reconstruction module is responsible for reconstructing the masked volumes at different levels. The cross-level alignment module aligns representations of volumes between volumes from adjacent levels, aiming to enforce anatomical similarity hierarchically.}}
	\label{fig:framework}
    \vspace{-17pt}
\end{figure*}

\subsection{Overall framework}\label{sec:framework}
The proposed MiM method's overall framework is presented in Fig.~\ref{fig:framework}, which comprises multi-level reconstruction modules and cross-level alignment modules. To pre-train the model with MiM, we first generate multi-level volumes from the input 3D medical images. Then, an input volume is cropped into non-overlapping patches, which are divided into unmasked patches and masked patches. The unmasked patches are transformed into a high-dimension feature space using a typical backbone (CNN~\cite{he2016deep} and transformer~\cite{dosovitskiy2020image}). The masked patches are used to generate the next level of masked volumes for multi-level learning. The goal is to restore the masked patches from different levels of masked volumes. In this paper, instead of a single level of masked volume~\cite{wang2023hard,wang2023masked}, we propose to ease this goal by multi-level of masked volume. We developed a $\mathcal{L}_{R}$ to supervise the final prediction. In addition, we further use a loss function $\mathcal{L}_C$ to align the shared semantics between the cross-level volumes, aiming to learn the global semantic content as well as local details.  Further details are presented in Section~\ref{sec:multi_reconstruction} and Section~\ref{sec:cross}.

\subsection{Multi-level reconstruction}\label{sec:multi_reconstruction}
\begin{figure}[!ht]
    \centering
	\vspace{-7pt}
	\includegraphics[width=1.0\linewidth]{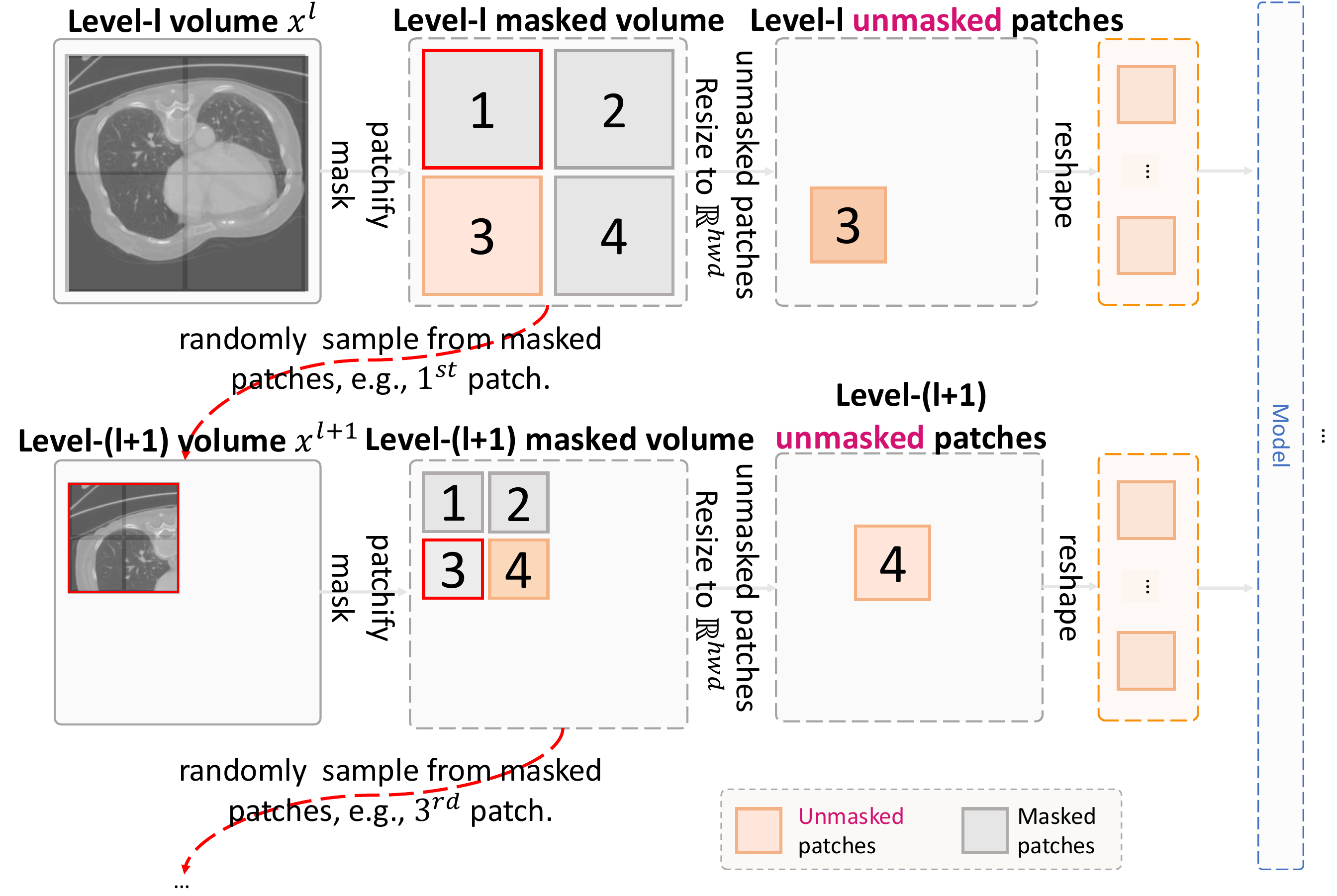}
	\vspace{-23pt}
	\caption{Illustration of the multi-level masked volume generation. Slices drawn from the 3D medical images are shown in the figure only for better illustration. Level-(l+1) volume $x^{l+1}$ is randomly sampled from the \textit{masked patches} of Level-l volume $x^l$ (\textit{i.e.,} patch with red box). }
	\label{fig:mim}
    \vspace{-5pt}
\end{figure}

\noindent\textbf{Generation of multi-level masked volumes.}
Given a volumetric image $x\in \mathbb{R}^{CHWD}$ (\textit{e.g.,} $C=1$ for CT), we aim to generate multi-level volumes $\{x^l \in \mathbb{R}^{C H^l W^l D^l}, l\in L\}$, which contains $L$ numbers of level, \textit{i.e.,} different granularity information from coarse to fine level. The generation of multi-level volume is illustrated in Fig.~\ref{fig:mim}. We start by cropping a large portion of sub-volume from $x$ as Level-1 volume $x^1\in \mathbb{R}^{C H^1 W^1 D^1}$. The level 1 volume $x^1$ is then patchified into $N$ non-overlapping visual tokens using a patch of $\{\frac{H^1}{6},\frac{W^1}{6},\frac{D^1}{6}\} $ as in ~\cite{chen2023masked,zhuang2023advancing}. As previous MAE-based methods~\cite{zhuang2023advancing}, we apply a high mask ratio $\mu$ (\textit{e.g.,} 60\%) to the $N$ non-overlapping tokens (\textit{e.g.,} 216), resulting in \ud{unmasked} tokens and \ud{masked} tokens. Each \ud{unmasked} token is resized at spatial dimension to $\mathbb{R}^{Chwd}$ before further \ud{processed by a linear projection} as previous MAE-based methods~\cite{chen2023masked}. The \ud{masked} tokens from Level-1 $x^1$ are considered as the next level volume, \textit{i.e.,} Level-2 volume $x^2\in \mathbb{R}^{C H^2 W^2 D^2}$. It's noted that the Level-2 volume $x^2$ is generated from the \textit{masked patches} of Level-1 volume $x^1$, instead of unmasked patches. The content of the target reconstruction at different levels shares regions, meaning that the volumes for Level-1 and Level-2 reconstruct overlapping regions but from different granularities, respectively. This can effectively capture the hierarchical structure of 3D medical images and improve representation learning~\cite{wang2023masked}. Ablation study of the reconstruction targets, \textit{i.e.,} the choices of masked patches or unmasked patches as next level volume, is presented in Section~\ref{sec:recontruction_target}. Since there exist many unmasked tokens, we randomly sample $\gamma$ times without replacement from them for computation efficiency. This generation process of multi-level volume is repeated until the last level of volume $x^L$ is generated. 

Thus, all \ud{unmasked} tokens from different levels are fed into the backbone to extract the high-dimension features $z$. Following~\cite{he2022masked,chen2023masked}, a lightweight decoder is used to project $z$ along with learnable \ud{masked} tokens into latent features $q$. For reconstruction, the decoder followed by a simple prediction head is used to reconstruct the \ud{masked} tokens $y$. 

\noindent\textbf{Single-level reconstruction.}~\label{sec:reconstruction}
Our reconstruction target is the pixel values of the \ud{masked} tokens for each level volume. With feature extracted from the backbone and the decoder with a prediction head (\textit{i.e.,} a linear layer), following previous MAE methods~\cite{chen2023masked}, we reshape the prediction results and reconstruction targets to one-dimensional vector, \textit{i.e.,} $\hat{y}\in \mathbb{R}^{{1\times C}}$ and $y\in \mathbb{R}^{{1\times C}}$, where $C$ is the number of dimensions. Specifically, we empirically set the $C$ to 768, as in ~\cite{chen2023masked,zhuang2023advancing}.

Then we compute the differences $d$ between the reconstruction target $y$ and the prediction result $\hat{y}$. Specifically, we use MSE distance~\cite{chen2023masked} to measure the difference $d_m$ for each masked token $m$ as follows:
\begin{equation}
    d_m =  \Vert y_m-\hat{y}_m \Vert_2, m\in M, M=\mu N,
\end{equation}
where $M$ denotes numbers of masked tokens.
To minimize the difference $d_m$, we define the reconstruction loss $\mathcal{L}_\mathcal{R}^{l}$ for each level of masked volumetric images $l\in L$ as follows:
\begin{equation}
	\label{eq:reconstructionLoss}
    \mathcal{L}_\mathcal{R}^{l}= \frac{1}{|M|}\sum_{m=1}^{|M|}d_m^l, l\in L,
\end{equation}
where $|M|$ represents the number of mask tokens.

\noindent\textbf{Loss function for multi-level reconstruction.}
To learn multi-granularity details, we apply \textit{single-level reconstruction} for each level of masked volumetric images. Thus, the multi-level reconstruction loss $\mathcal{L}_\mathcal{R}$ can be formulated as follows:
\begin{equation}\label{eq:multi_level_reconstruction}
    \mathcal{L}_\mathcal{R}= \frac{1}{L}\sum_{l=1}^{L}\mathcal{L}_\mathcal{R}^{l}
\end{equation}

\subsection{Cross-level alignment}\label{sec:cross}
\begin{figure}[!ht]
    \centering
	\vspace{-5pt}
	\includegraphics[width=1.0\linewidth]{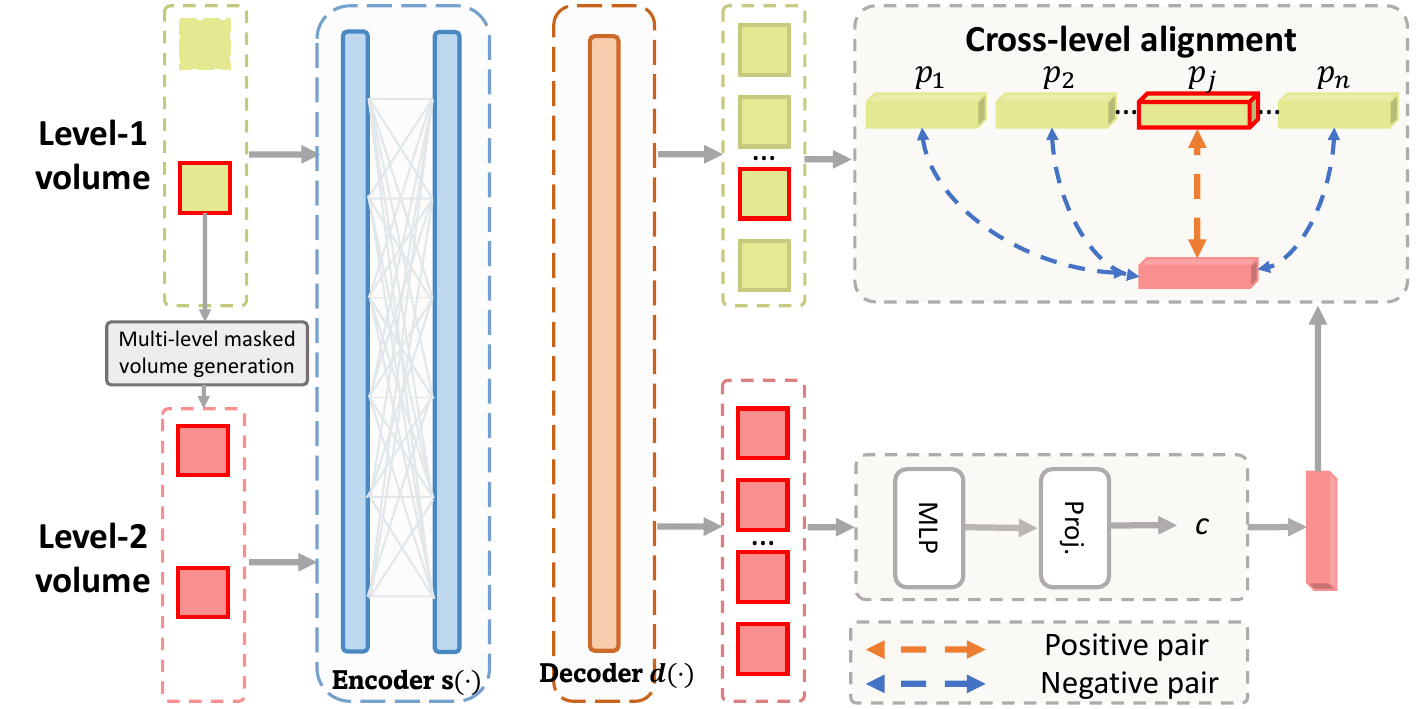}
	\vspace{-20pt}
	\caption{Illustration of cross-level alignment module. We set level $L$ to 2 for better illustration. 
 }
	\label{fig:cross_level_alignment}
    \vspace{-10pt}
\end{figure}

The alignment between the shared semantic patches of cross-level volumes from fine-to-coarse can enforce anatomical similarity in a hierarchical manner. Fig.~\ref{fig:cross_level_alignment} illustrates the process of cross-level alignment.  Since we generate the finer level volumes from the coarser level volumes (\textit{e.g.,} Level-2 volumes generated from masked patches of the Level-1 volumes), these volumes must share semantic context, which can be regarded as positive pairs. In contrast, the non-overlap patches in the coarser volumes (\textit{e.g.,} the rest patches in the Level-1 volumes) are considered as negative patches. To enlarge the high-dimensional feature consistency for shared anatomical structure patches (\textit{i.e., positive pairs}) and discrepancy between non-overlap patches (\textit{i.e., negative pairs}), we apply contrastive learning~\cite{chen2020simple} to the context vector $c$ and patches $p$ . Specially, we reshape the context vector and patches its coarser volumes to one-dimension vector $c\in \mathbb{R}^{1\times D}$ and $\{p_i\in \mathbb{R}^{1\times D}, i\in N\}$, where $D$ is the number of dimensions, empirically set $D$ to 2048, as in ~\cite{caron2021emerging}.

First, given the context vector $c_i$ and the patches $p$, we compute the cosine similarity $s_{ij}$ between the context vector $c_i$ and the patches $p_j$ as follows:

\begin{equation}\label{eq:cossim}
    s_{ij} = cosSim(c_i, p_j) = \frac{c_i\cdot p_j}{\|c_i\|\ \|p_j\|}, i\in M, j\in N.
\end{equation}

We aim to maximize cosine similarity between the context vector $c_i$ and the positive patches $p$ and minimize  cosine similarity between the context vector $c_i$ and the negative patches $p_j$ from the rest patches of the volumes. Thus, we apply contrastive loss~\cite{chen2020simple} to implement this goal as follows,
\begin{equation}\label{eq:nce}
    \ell_{ij} = -\textrm{log}\frac{\textrm{exp}\ (s_{ij}/\tau)}{\sum_{k=1}^{N}\mathbbm{1}_{k\neq i}\ \textrm{exp}(s_{ik}/\tau)}, i\in M, j\in N, M\in \mu N.
\end{equation}

Thus, the cross-level alignment loss $\mathcal{L}_\mathcal{C}^{l, l+1}$ between adjacent level $l$ and $l+1$ volumes is computed as follows:
\begin{equation}\label{eq:cross_loss}
    \mathcal{L}_{\mathcal{C}}^{l, l+1} = \frac{1}{M\cdot N}\sum_{i=1}^{M}\sum_{j=1}^{N}l_{ij}, l\in L-1, M\in \mu N.
\end{equation}
The cross-level alignment loss is computed as the sum of the negative log-likelihood of the cosine similarity between the context vector and the rest patches from the coarser volumes.

The formulation of cross-level alignment $\mathcal{L}_{\mathcal{C}}$ is defined as the average of the cross-level alignment loss between adjacent level volumes, as shown in Eq.~\ref{eq:overall_cross_loss},
\begin{equation}\label{eq:overall_cross_loss}
    \mathcal{L}_{\mathcal{C}} = \frac{1}{L-1}\sum_{l=1}^{L-1}\mathcal{L}_{\mathcal{C}}^{l, l+1}
\end{equation}
By minimizing the overall cross-level alignment loss, we can enforce anatomical similarity hierarchically.

\subsection{Backbone}\label{sec:backbone}
\begin{figure}[!h]
	\centering
	\includegraphics[width=1\linewidth]{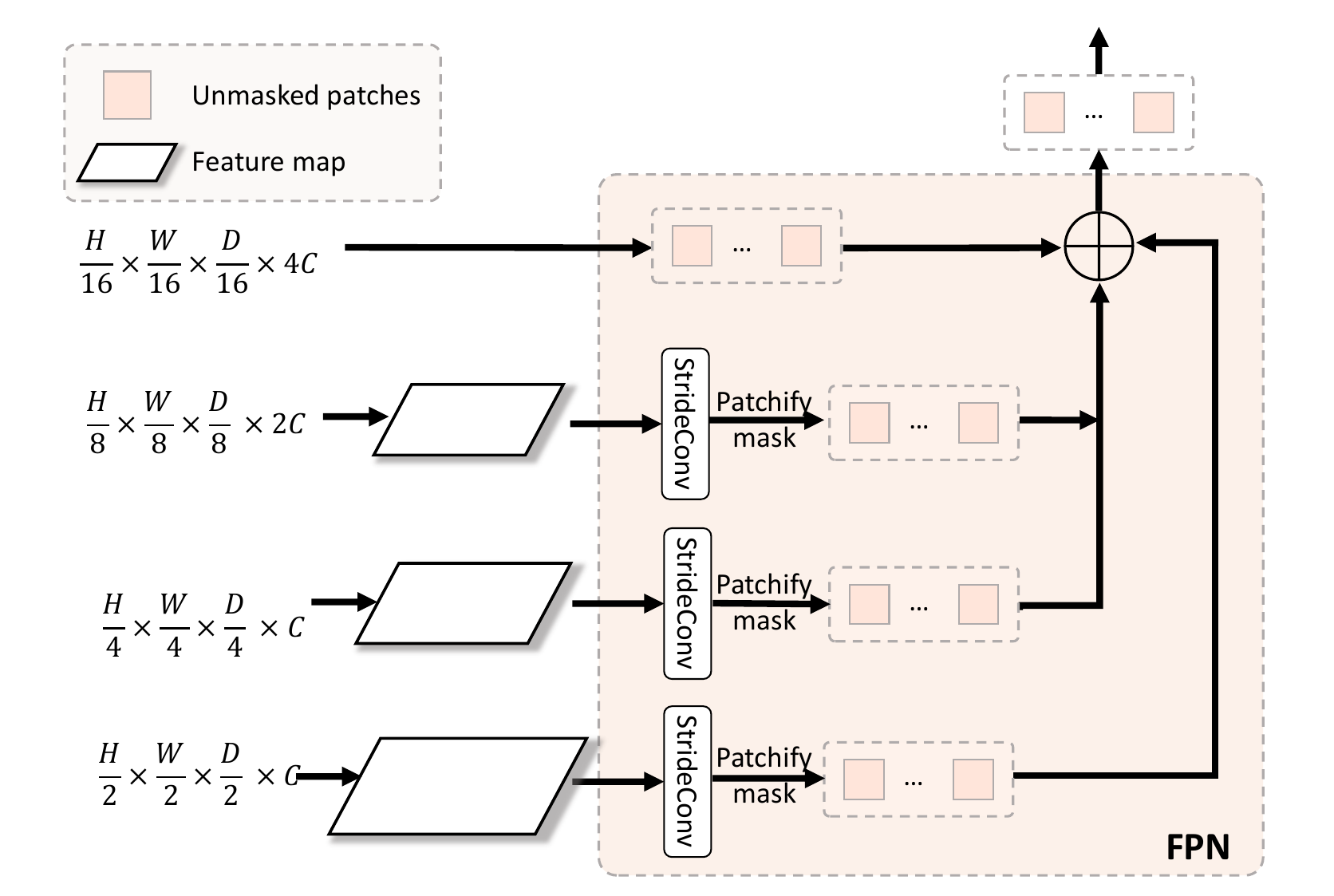}
        \vspace{-20pt}
	\caption{\ud{Architecture of the FPN adapted from MCMAE~\cite{gao2022convmae}, illustrating multi-scale feature extraction and fusion through hierarchical feature maps.}}
	\label{fig:fpn}
\end{figure}

\ud{While previous hybrid transformers like Swin Transformer~\cite{hatamizadeh2021swin} generate pyramid features, they process all tokens during Masked Image Modeling pre-training, leading to computational inefficiency~\cite{gao2022convmae}. We address this limitation by extending MCMAE's backbone~\cite{gao2022convmae} to 3D medical images through an additional depth dimension, allowing our model to process only unmasked tokens in transformer layers. This optimization significantly improves computational efficiency and scalability, as validated in Section~\ref{sec:ablation_backbone}. 
As illustrated in Fig~\ref{fig:fpn}, our FPN architecture adopts MCMAE's~\cite{gao2022convmae} hierarchical design, processing features at four scales (H/2×W/2 to H/16×W/16) with channel dimensions (C to 4C). Each scale utilizes StrideConv~\cite{gao2022convmae} downsampling and is then patchified to patches, followed by masking for unmasked patch generation. The bottom-up pathway integrates features through lateral connections and summations, producing comprehensive multi-scale representations while maintaining computational efficiency.}

\noindent\textbf{Overall objective function.}\label{sec:objective}
Our MiM introduces a hierarchically designed approach to 3D medical image representation learning through multi-level reconstruction $\mathcal{L}_R$ and cross-level alignment $\mathcal{L}_C$. We empirically set the level of masked volume to $L=3$ since the three-level masked volume can provide a good balance between the representation learning and computational efficiency. Ablation study of $L$ is on the Section~\ref{sec:ablation}. Specifically, the multi-level reconstruction loss in Eq.~\ref{eq:multi_level_reconstruction} can be further expanded as follows:

\begin{equation}\label{eq:full_multi_level_reconstruction}
    \mathcal{L}_\mathcal{R}= \mathcal{L}_\mathcal{R}^{1} + \mathcal{L}_\mathcal{R}^{2} + \mathcal{L}_\mathcal{R}^{3}.
\end{equation}

Then, since cross-level alignment loss applied between adjacent level volumes, Eq.~\ref{eq:cross_loss} expands as follows:
\begin{equation}
    \mathcal{L}_\mathcal{C}= \mathcal{L}_\mathcal{C}^{1,2} + \mathcal{L}_\mathcal{C}^{2,3}.
\end{equation}

Thus, the total loss function $\mathcal{L}$ is the combination of these two losses, as shown in Eq.~\ref{eq:loss},
\begin{equation}\label{eq:loss}
    \mathcal{L}= \mathcal{L}_\mathcal{R} + \alpha\mathcal{L}_\mathcal{C},
\end{equation}
where the hyper-parameters $\alpha$ are used to balance the relative contributions of these two kinds of losses. We empirically set the $\alpha$ to 0.1 based on our experiment results. The ablation study of hyperparameter $\alpha$ presents in Section~\ref{sec:hyperparameter}.

\section{Experiments}\label{sec:experiments}

This section will commence by introducing the datasets and evaluation metrics. Then, we will elaborate on the implementation details of our MiM. Lastly, we will present the experimental results of MiM in comparison to existing methods, along with an analysis of our proposed approach.

\subsection{Datasets and Evaluation}
\noindent\textbf{Pre-training datasets.}\label{sec:pretrainingDataset}
To conduct a fair comparison with previous works~\cite{chen2023masked,tang2022self,he2023geometric,wang2023swinmm,xie2022unimiss,zhou2021models,zhou2021preservational,zhou2023unified,zhuang2023advancing}, we also carried out pre-training experiments on two public datasets, \textit{i.e.,} BTCV~\cite{landman2015miccai} and TCIA Covid19~\cite{clark2013cancer}, and combined them to form a new dataset named \textbf{\textit{1k}}. Additionally, to explore the scaling ability of our proposed method compared to previous state-of-the-art methods~\cite{chen2023masked,tang2022self}, we collected eight publicly accessible 3D medical image datasets consisting of 10,502 CT scans to establish our pre-training datasets, which we named \textbf{\textit{10k}}. It is important to note that \textbf{\textit{10k}} is only used for exploring the scaling ability of our proposed method, while we mainly focus on the \textbf{\textit{1k}} dataset for fair comparison and analysis of our proposed method. Table~\ref{tab:pretraining_dataset} provides a summary of the sources of each collected dataset.

\begin{table}[!ht]
    \centering
    \vspace{-15pt}
    \caption{The details of each dataset in our pre-training datasets.}\label{tab:pretraining_dataset}
    \vspace{-5pt}
    \resizebox{0.90\linewidth}{!}{
    \begin{tabular}{llccr}
    \whline
        \multirow{2}{*}{Datasets} & \multirow{2}{*}{Region of Interest} &  \multicolumn{2}{c}{\tabincell{c}{Pre-training}} & \multirow{2}{*}{\#Samples}\\
        \cline{3-4}
        & & \textbf{1k} & \textbf{10k} & \\
        \whline
        BTCV~\cite{landman2015miccai}                          & Abdomen        & \CheckmarkBold & \CheckmarkBold & 24\\
        TCIA Covid19\ud{~\cite{clark2013cancer}}                           & Chest          & \CheckmarkBold & \CheckmarkBold & 722\\    
        LUNA16~\cite{harmon2020artificial}                     & Chest          &                & \CheckmarkBold & 843\\
        STOIC 2021~\cite{revel2021study}                       & Chest          &                & \CheckmarkBold & 2,000\\
        FLARE23~\ud{\cite{ma2024automatic}}                                                & Abdomen        &                & \CheckmarkBold & 4,000\\
        LiDC~\cite{armato2011lung}                             & Chest          &                & \CheckmarkBold & 589\\
        HNSCC~\cite{grossberg2018imaging}                      & Head/Neck      &                & \CheckmarkBold & 1,071\\
        TotalSegmentator~\cite{wasserthal2022totalsegmentator} & \tabincell{l}{Head/Neck/Chest/leg\\Abdomen/pelvis/feet } & & \CheckmarkBold & 1,203\\
        \rowcolor{myred}{\it\bf Total} & & & & 10,502\\
        \whline
    \vspace{-20pt}
    \end{tabular}
    }

\end{table}

\noindent\textbf{Downstream datasets.} We conduct experiments on twelve public datasets, \textit{i.e,} BTCV~\cite{landman2015miccai}, MM-WHS~\ud{\cite{zhuang2018multivariate}}, Spleen~\cite{simpson2019large}, Flare22~\cite{ma2023unleashing}, Amos22~\cite{ji2022amos}, MSD Task03~\cite{simpson2019large}, MSD Task06~\cite{simpson2019large}, MSD Task07~\cite{simpson2019large}, MSD Task08~\cite{simpson2019large}, MSD Task10~\cite{simpson2019large}, BrasTS 21~\cite{simpson2019large}, and CC-CCII~\cite{zhang2020clinically}.  These datasets include segmentation and classification tasks, with the first ten datasets used for organ segmentation, the eleventh dataset used for lesion segmentation, the twelfth dataset used for tumor segmentation, and the last dataset used for disease classification. 
\ud{For BTCV~\cite{landman2015miccai} dataset, we adhere strictly to the data splits defined in prior studies~\cite{tang2022self,zhuang2023advancing,VoCo}, which comprise only training and validation sets. The training split is utilized for both pre-training and fine-tuning, while the validation split is excluded from pre-training and reserved solely for evaluation. All other datasets are unseen during pre-training.} Furthermore, to evaluate the cross-modality generalization ability, we transferred the model pre-trained on CT scans to the MRI dataset BrasTS 21~\cite{simpson2019large}. We adopted consistent settings as previous work~\cite{chen2023masked,tang2022self,zhuang2023advancing}.

\noindent\textbf{Evaluation metrics.}\label{sec:evaluation}
Following~\cite{tang2022self,zhuang2023advancing}, we utilized Dice Similarity Coefficient (DSC) and Normalized Surface Distance (NSD) to evaluate the segmentation tasks. We then utilized Accuracy (ACC) and the Area Under the Curve (AUC) to evaluate the disease classification tasks.
\begin{table}[!ht]
    \caption{Pre-training settings.}
    \centering
    \label{tab:pretrainerdsetting}
    \vspace{-5pt}
    \resizebox{0.87\linewidth}{!}{
    \begin{tabular}{lc}
        \whline
        \multicolumn{2}{c}{Pre-training settings}\\
    \hline
        Steps                                            & 45K\\
        Optimizer                                        & AdamW\\
        Learning rate (LR)                               & 1e-4\\
        LR scheduler                                     & cosine annealing schedule\\
        Warmup step                                      & 100\\
        Regularization weight                            & 1e-2\\
        Batch size                                       & 256\\
        MiM Level-1 \ud{$H^1 W^1 D^1$}                        & \ud{$384,384,192$}\\
        MiM Level-2 \ud{$H^2 W^2 D^2$}                        & \ud{$96,96,96$}\\
        MiM Level-3 \ud{$H^3 W^3 D^2$}                        & \ud{$16,16,16$}\\
        MiM Resize after crop \ud{$h w d$}                    & \ud{$96,96,96$}\\
        MiM mask ratio $\tau$                                & 0.6\\
        MiM sampling times on next level $\gamma$                   & 4\\
        \whline
    \end{tabular}
    }
    \vspace{-11pt}
\end{table}

\begin{table*}[htb]
    \centering
    \vspace{-20pt}
    \caption{Experiment results on BTCV dataset~\cite{landman2015miccai} across thirteen organs.
    The best results are highlighted in red.
    The term `From Scratch' signifies the supervised baseline without self-supervised pre-training.
    $\dag$ denotes that we re-implement the approach. Most results are drawn from~\cite{chen2023masked,zhang2023dive,zhuang2023advancing} or their papers.}
    \label{tab:btcv_benchmark}
    \vspace{-7pt}
    \resizebox{1\textwidth}{!}{
    \begin{tabular}{lcccccccccccccccc}
		\whline
\multirow{2}{*}{Method} & \multirow{2}{*}{\makecell[c]{Pre-traning \\dataset}} & \multirow{2}{*}{\ud{Network}} & \multicolumn{13}{c}{\textbf{Dice Score(\%)}} & \multirow{2}{*}{\textbf{Avg}}\\
\cline{4-16} & & & Spl & RKid & LKid & Gall & Eso & Liv & Sto & Aor & IVC & Veins & Pan & RAG & LAG & \\
\midrule
  \textit{\textbf{From Scratch}}\\
UNETR~\cite{hatamizadeh2022unetr}            & -   & \ud{-} & 93.02 & 94.13 & 94.12 & 66.99 & 70.87 & 96.11 & 77.27 & 89.22 & 82.10 & 70.16 & 76.65 & 65.32 & 59.21 & 79.82\\
\rowcolor{mygray}
Swin-UNETR~\cite{hatamizadeh2021swin}        & -   & \ud{-} & 94.06 & 93.54 & 93.80 & 65.51 & 74.60 & 97.09 & 75.94 & 91.80 & 82.36 & 73.63 & 75.19 & 68.00 & 61.11 & 80.53\\
\hline
\textit{\textbf{General SSL}}\\
SimCLR~\cite{chen2020simple}                 & 1k  & \ud{UNETR} & 92.79 & 93.04 & 91.41 & 49.65 & 50.99 & 98.49 & 77.92 & 85.56 & 80.58 & 64.37 & 67.16 & 59.04 & 48.99 & 73.85\\
MoCov3~\cite{chenempirical}                  & 1k  & \ud{UNETR} & 91.96 & 92.85 & 92.42 & 68.25 & 72.77 & 94.91 & 78.82 & 88.21 & 81.59 & 71.15 & 75.76 & 66.48 & 58.81 & 79.54\\
DINO$\dag$~\cite{caron2021emerging}          & 1k  & \ud{UNETR} & 93.64 & 92.95 & 92.77 & 74.70 & 71.87 & 96.47 & 77.85 & 89.49 & 83.30 & 72.12 & 78.41 & 67.26 & 63.88 & 81.22\\
localMIM$\dag$~\cite{wang2023masked}         & 1k  & \ud{UNETR}  & 95.31 & 94.16 & 94.17 & 74.52 & 73.69 & 96.57 & 82.21 & 89.92 & 84.67 & 72.12 & 76.89 & 67.68 & 62.29 & 81.96  \\
HPM$\dag$~\cite{wang2023hard}                & 1k  & \ud{UNETR}  & 94.47 & 93.46 & 93.86 & 75.62 & 74.07 & 96.11 & 80.92 & 90.01 & 84.42 & 71.25 & 79.29 & 67.34 & 64.40 & 82.03   \\
SimMIM~\cite{xie2022simmim}                  & 1k  & \ud{Swin-UNETR} & 95.51 & 93.61 & 93.49 & 67.91 & 73.50 & 96.46 & 81.15 & 89.78 & 84.86 & 72.45 & 75.70 & 66.89 & 64.46 & 81.41\\
MCMAE$\dag$~\cite{gao2022convmae}            & 1k & \ud{Swin-UNETR} & 94.60 & 94.08 & 93.87 & 62.66 & 75.13 & 96.26 & 82.08 & 90.27 & 85.68 & 75.99 & 81.18 & 68.78 & 64.68 & 82.20\\
\hline
\textit{\textbf{3D Medical SSL}}\\
ROT~\cite{taleb20203d}                       & 1k & \ud{3D UNet} & 91.75 & 93.13 & 91.62 & 65.09 & 76.55 & 94.21 & 86.16 & 89.74 & 83.08 & 71.13 & 81.55 & 67.90 & 63.72 & 81.20\\
ModelGen~\cite{zhou2021models}               & 1k & \ud{3D UNet} & 91.99 & 93.52 & 91.81 & 65.11 & \bb{76.14} & 95.98 & \rr{86.88} & 89.29 & 83.59 & 71.79 & 81.62 & 67.97 & 63.18 & 81.45\\
PCRLv1$\dag$~\cite{zhou2021preservational}   & 1k & \ud{3D UNet} & 94.44 & 92.50 & 92.75 & 56.46 & 74.95 & 96.62 & 81.64 & 89.86 & 87.12 & 72.78 & 75.24 & 69.73 & \bb{68.18} & 81.30\\
PCRLv2$\dag$~\cite{zhou2023unified}          & 1k & \ud{3D UNet} & 95.50 & 91.43 & 89.52 & \rr{76.15} & 73.54 & \bb{97.28} & 79.64 & 90.16 & 84.17 & \rr{75.20} & 78.71 & 68.74 & 62.93 & 81.74\\
UniMiss$\dag$~\cite{xie2022unimiss}          & 1k & \ud{MiT} & 95.24 & 93.74 & 93.78 & 72.69 & 73.61 & 96.23 & 83.08 & 88.50 & 83.31 & 71.89 & 78.60 & 66.22 & 68.03 & 82.05\\
GL-MAE~\cite{zhuang2023advancing}           & 1k  & \ud{UNETR} & 94.54 & \bb{94.39} & \bb{94.37} & 73.19 & 74.93 & 96.51 & 83.49 & 89.74 & 83.11 & 70.80 & 75.71 & 69.39 & 63.12 & 82.33\\
MAE3D~\cite{chen2023masked}                 & 1k  & \ud{UNETR} & 95.81 & 94.38 & 94.48 & 69.96 & 76.85 & 96.69 & 80.44 & 90.33 & 84.33 & 73.65 & 80.11 & 68.65 & 64.44 & 82.40\\
Rubik++~\cite{zhu2020rubik}                  & 1k & \ud{Swin-UNETR} & \rr{96.21} & 90.41 & 89.33 & 75.22 & 72.64 & \rr{97.44} & 79.25 & 89.65 & 83.76 & 74.74 & 78.35 & 67.14 & 61.97 & 81.38\\
SwinMM~\cite{wang2023swinmm}                 & 1k & \ud{Swin-UNETR} & 94.33 & 94.18 & 94.16 & 72.97 & 74.75 & 96.37 & 83.23 & 89.56 & 82.91 & 70.65 & 75.52 & 69.17 & 62.90 & 81.81\\
Adam$\dag$~\cite{hosseinzadeh2023towards}    & 1k & \ud{Swin-UNETR} & 94.16 & 93.65 & 93.43 & 66.14 & 71.28 & 96.18 & 76.93 & 89.83 & 85.35 & 71.16 & 80.37 & 63.97 & 60.83 & 80.45\\
GVSL$\dag$~\cite{he2023geometric}            & 1k & \ud{Swin-UNETR} & 95.27 & 91.22 & 92.25 & 72.69 & 73.56 & 96.44 & 82.40 & 88.90 & 84.22 & 70.84 & 76.42 & 67.48 & 63.25 & 81.87\\
Swin-UNETR$\dag$~\cite{tang2022self}        & 1k  & \ud{Swin-UNETR} & 95.91 & 94.48 & 94.42 & 69.57 & 76.47 & 96.94 & 78.50 & 90.31 & 85.77 & 75.12 & 81.33 & 67.37 & 64.92 & 82.58\\
\rowcolor{myred}
\textbf{MiM}     & 1k                       & \ud{Swin-UNETR} & \bb{96.05} & \rr{94.58} & \rr{94.53} & \bb{75.73} & \rr{77.36} & \bb{97.03} & \bb{83.23} & \rr{90.37} & \rr{87.64} & \bb{74.97} & \rr{82.62} & \rr{71.02} & \rr{68.80} & \rr{84.46}\\
\cdashline{1-15}
MAE3D$\dag$~\cite{chen2023masked}          & 10k & \ud{UNETR} & \bb{95.91} & 94.28 & 94.26 & 73.82 & 75.35 & 97.07 & 83.53 & \rr{91.12} & 86.74 & 75.13 & \bb{83.33} & 68.43 & 66.40 & \bb{83.52}\\
Swin-UNETR$\dag$~\cite{tang2022self}       & 10k & \ud{Swin-UNETR} & 95.20 & 94.30 & 94.15 & \bb{76.53} & 74.02 & 96.61 & 82.25 & 89.84 & 84.20 & 74.23 & 82.53 & \bb{70.74} & 67.49 & 83.20\\
\rowcolor{myred}
\textbf{MiM}     & 10k & \ud{Swin-UNETR} & \rr{96.37} & \rr{94.83} & \rr{94.75} & \rr{81.02} & \rr{80.08} & \rr{97.12} & \rr{85.30} & \bb{90.36} & \rr{87.66} & \rr{75.99} & \rr{84.41} & \rr{71.94} & \rr{69.64} & \rr{85.41}\\
\whline
    \end{tabular}
    }\\
\textsuperscript{*}{\ud{\raggedleft{\scriptsize{Note: Spl: spleen, RKid: right kidney, LKid: left kidney, Gall: gallbladder, Eso: esophagus, Liv: liver, Sto: stomach, Aor: aorta, IVC: inferior vena cava, Veins: portal and splenic veins, Pan: pancreas, AG: left and right adrenal glands.}}}}

\end{table*}

\begin{figure*}[!th]
     \centering
      \vspace{-10pt}
      \includegraphics[width=0.7\linewidth]{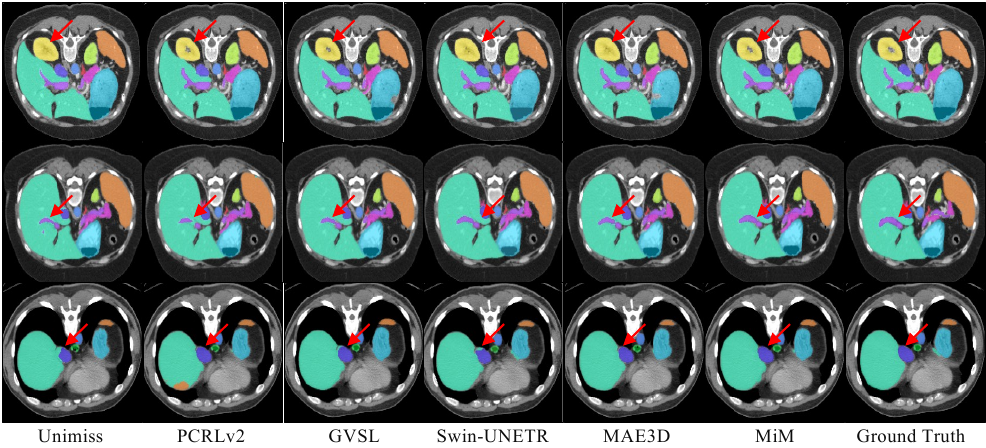}
      \vspace{-10pt}
      \caption{Visualization of segmentation results on BTCV validation dataset~\cite{landman2015miccai}. We compared MiM with Unimiss~\cite{xie2022unimiss}, PCRLv2~\cite{zhou2023unified}, GVSL~\cite{he2023geometric}, Swin-UNETR~\cite{tang2022self} and MAE3D~\cite{chen2023masked}.}
      \vspace{-20pt}
      \label{fig:SegVisualizatoin}
  \end{figure*}

\subsection{Implementation Details}
During pre-training, we followed the settings of previous works~\cite{chen2023masked,tang2022self} and provided the details of our MiM pre-training settings in Table~\ref{tab:pretrainerdsetting}. \ud{Specifically, the level-1 volume was randomly cropped from the entire CT volumes.} We used the backbone of ~\cite{gao2022convmae} as the encoder for efficient token processing. As previous SSL~\cite{caron2021emerging,zhuang2023advancing}, the prediction head and projection head are implemented with MLP layers for aligning the dimension. During fine-tuning, we used Swin-UNETR~\cite{hatamizadeh2021swin} for fine-tuning on segmentation tasks and Swin-ViT for fine-tuning on classification tasks, as per the settings of previous works~\cite{tang2022self,zhuang2023advancing}. \ud{Specifically, for segmentation task, we discarded the decoder and only used the backbone in the finetuning period. For classification tasks,we strictly adhered to the methodology used in prior work in general computer vision~\cite{caron2021emerging,he2022masked} and 3D medical imaging~\cite{zhuang2023advancing}, utilizing only the features from the final layer for pre-training. While incorporating multiscale features could enhance classification performance~\cite{liu2021swin,fan2021multiscale,du2022swinpa}, we opted not to follow this approach to ensure a fair comparison with other methods. Therefore, we use only the final layer’s features, combined with a global average pooling (GAP) layer and a simple MLP classifier, to predict the category.} We initialized the encoder of the network with learned parameters in the pre-training process and fine-tuned the overall network.
For inference on these datasets, we applied a \ud{sliding window} inference with overlapping to enable a fair comparison with previous works~\cite{tang2022self}. 
It's important to note that for evaluating the pure effectiveness of our proposed method, we did not use any foundation model or post-processing techniques~\cite{jiang2023anatomical,liu2023clip}.

\noindent\textbf{Comparison methods.} We compared our MiM method with both General SSL and Medical SSL methods. First, we compare the typical SSL methods MoCov3~\cite{chenempirical} and MAE~\cite{chen2023masked,he2022masked}, since they represent the two mainstream SSL paradigms. We also report the results of SimCLR~\cite{chen2020simple} according to ~\cite{chen2023masked,zhuang2023advancing}. We further evaluate the performance of SimMIM~\cite{xie2022simmim}, HPM~\cite{wang2023swinmm}, localMIM~\cite{chen2023masked} and MCMAE~\cite{gao2022convmae}, since they are related to our advanced hybrid MAE.
We also compare with Adam~\cite{hosseinzadeh2023towards} due to related to our hierarchical design of multi-granularity. Additionally, we compared with most existing SOTA medical SSL methods. 
\ud{Following common practices in SSL for natural images~\cite{he2019moco,chen2020simple,he2022masked,chenempirical,caron2021emerging} and 3D medical images~\cite{tang2022self,chen2023masked,zhuang2023advancing,zhou2021preservational,zhou2023unified}, we performed a single round of pre-training and fine-tuning for all methods to obtain results.}
\begin{table*}[!ht]
    \centering
    \vspace{-10pt}
    \caption{Experiment results on BTCV~\cite{landman2015miccai}, MM-WHS~\cite{zhuang2018multivariate}, MSD Spleen~\ud{\cite{simpson2019large}}, Amos22~\cite{ji2022amos} and Flare22~\cite{ma2023unleashing} with varying ratios of training datasets. The evaluation was conducted using the DSC(\%) metric.
    $\dag$ denotes we re-implement the approach and report the results. Most results are drawn from~\cite{chen2023masked,zhuang2023advancing}.}
      \vspace{-8pt}
    \resizebox{0.98\linewidth}{!}{
    \begin{tabular}{lc|ccc|ccc|ccc|ccc|ccc|c}
    \whline
        \multirow{2}{*}{Method} & \multirow{2}{*}{\makecell[c]{Pre-traning \\dataset}} &\multicolumn{3}{c|}{BTCV} & \multicolumn{3}{c|}{MM-WHS} & \multicolumn{3}{c|}{MSD Spleen}  & \multicolumn{3}{c|}{Amos22} & \multicolumn{3}{c|}{Flare22} & \multirow{2}{*}{\textbf{Avg}}\\
        \cline{3-17}
        & & 25\% & 50\% & 100\% & 25\% & 50\% & 100\% & 25\% & 50\% & 100\% & 25\% & 50\% & 100\% & 25\% & 50\% & 100\% & \\
        \hline
\textbf{\textit{From scratch}}\\
UNETR~\cite{hatamizadeh2022unetr}                & - & 58.99 & 75.20 & 79.82 & 78.77 & 85.31 & 85.85 & 89.67 & 92.23 & 94.20 & 73.53 & 81.46 & 84.34 & 75.32 & 76.29 & 76.94 & 80.53\\
\rowcolor{mygray}
Swin-UNETR~\cite{hatamizadeh2021swin}           & - & 63.08 & 75.38 & 80.53 & 79.59 & 85.65 & 86.11 & 90.08 & 93.29 & 94.90 & 76.03 & 82.60 & 84.85 & 77.43 & 80.50 & 83.08 & 82.21\\
\hline
\textbf{\textit{General SSL}}\\
SimCLR~\cite{chen2020simple}               & 1k & 59.03 & 75.31 & 79.85 & 78.88 & 85.52 & 86.00 & 89.88 & 92.52 & 94.11 & 73.62 & 81.65 & 84.66 & 75.34 & 76.33 & 77.03 & 80.65\\
MoCov3$\dag$~\cite{chenempirical}         & 1k & 49.55 & 75.65 & 79.54 & 79.50 & 86.13 & 84.16 & 92.12 & 93.56 & 94.23 & 76.11 & 82.70 & 84.95 & 77.62 & 80.91 & 83.22 & 81.33\\
DINO$\dag$~\cite{caron2021emerging}           & 1k & 59.68 & 75.53 & 81.22 & 78.80 & 86.31 & 89.80 & 93.58 & 93.84 & 95.79 & 74.40 & 84.43 & 87.24 & 78.42 & 82.01 & 85.99 & 83.14\\
SimMIM$\dag$~\cite{xie2022simmim}         & 1k & 62.01 & 75.98 & 81.41 & 79.71 & 86.38 & 90.22 & 93.87 & 93.48 & 94.94 & 79.57 & 86.97 & 89.74 & 78.01 & 82.49 & 86.05 & 84.01\\
localMIM$\dag$~\cite{wang2023masked}       & 1k & 62.26 & 77.43 & 81.96 & 82.21 & 87.15 & 89.99 & 91.80 & 92.48 & 93.37 & 79.84 & 85.72 & 88.09 & 78.43 & 82.42 & 86.48 & 83.98\\
HPM$\dag$~\cite{wang2023hard}            & 1k & 62.95 & 76.32 & 82.03 & 82.34 & 87.64 & 89.86 & 93.78 & 93.11 & 93.86 & 79.89 & 85.70 & 88.07 & 78.20 & 82.42 & 86.15 & 84.16\\
MCMAE$\dag$~\cite{gao2022convmae}          & 1k & 64.70 & 78.01 & 82.20 & 85.24 & 88.74 & 90.29 & 94.92 & 95.88 & 96.04 & 75.82 & 84.76 & 86.63 & 78.97 & 80.79 & 84.31 & 84.49\\
\hline
\textbf{\textit{3D Medical SSL}}\\
ROT~\cite{taleb20203d}                  & 1k & 53.43 & 74.44 & 81.20 & 80.12 & 86.98 & 87.12 & 90.99 & 95.01 & 95.45 & 76.88 & 86.01 & 87.12 & 77.11 & 81.16 & 84.51 & 82.50\\
Rubik++~\cite{zhu2020rubik}              & 1k & 53.00 & 74.74 & 81.38 & 80.88 & 87.18 & 88.02 & 91.91 & 95.22 & 95.73 & 77.43 & 86.91 & 88.20 & 78.43 & 82.85 & 86.49 & 83.22\\
ModelGen~\cite{zhou2021models}             & 1k & 54.18 & 61.26 & 81.45 & 80.33 & 86.26 & 89.69 & 92.01 & 95.22 & 95.73 & 77.49 & 86.20 & 88.29 & 78.17 & 82.37 & 84.02 & 82.18\\
PCRLv1$\dag$~\cite{zhou2021preservational}         & 1k & 55.08 & 70.92 & 81.30 & 83.34 & 87.59 & 90.32 & 89.69 & 92.01 & 95.10 & 79.46 & 86.24 & 86.02 & 78.44 & 84.14 & 87.34 & 83.13\\
PCRLv2$\dag$~\cite{zhou2023unified}         & 1k & 58.67 & 71.40 & 81.32 & 84.51 & 88.13 & 90.36 & 90.32 & 95.16 & 95.46 & 77.32 & 84.92 & 88.79 & 79.10 & 85.37 & 87.45 & 83.89\\
SwinMM$\dag$~\cite{wang2023swinmm}         & 1k & 59.40 & 73.66 & 81.81 & 84.61 & 88.32 & 90.40 & 90.36 & 95.23 & 95.55 & 77.33 & 84.95 & 88.88 & 79.26 & 85.62 & 87.56 & 84.20\\
Adam$\dag$~\cite{hosseinzadeh2023towards}           & 1k & 64.12 & 73.66 & 80.45 & 83.12 & 87.26 & 88.82 & 88.73 & 92.09 & 86.90 & 59.89 & 73.89 & 88.87 & 79.54 & 84.43 & 87.24 & 81.27\\
GVSL$\dag$~\cite{he2023geometric}           & 1k & 52.03 & 74.23 & 81.87 & 85.24 & 88.00 & 90.22 & 95.36 & 96.20 & 96.57 & 78.72 & 87.41 & 89.92 & 81.89 & 85.38 & 87.42 & 84.70\\
Unimiss$\dag$~\cite{xie2022unimiss}        & 1k & 65.96 & 77.15 & 82.05 & 83.64 & 88.49 & 90.37 & 93.30 & 94.74 & 95.10 & 77.03 & 85.08 & 88.23 & 80.37 & 85.21 & 87.66 & 84.96\\
GL-MAE~\cite{zhuang2023advancing}               & 1k & 66.44 & 76.37 & 82.33 & 76.16 & 87.72 & 88.88 & 93.65 & 94.36 & 95.72 & 74.65 & 84.65 & 87.40 & 80.20 & 85.03 & 87.26 & 84.05\\
Swin-UNETR$\dag$~\cite{tang2022self}     & 1k & 62.12 & 76.13 & 82.58 & 85.33 & 87.50 & 90.32 & 95.13 & 95.00 & 95.02 & 78.41 & 87.49 & 89.83 & 83.21 & 85.72 & 87.17 & 85.40\\
MAE3D$\dag$~\cite{chen2023masked}          & 1k & 63.66 & 78.57 & 82.48 & 85.57 & 88.62 & 90.03 & 95.12 & 95.11 & 95.50 & 77.63 & 87.08 & 89.71 & 83.31 & 85.85 & 87.31 & 85.70\\
\rowcolor{myred}
\textbf{MiM}         & 1k & \rr{70.71} & \rr{81.22} & \rr{84.66} & \rr{86.52} & \rr{88.99} & \rr{91.04} & \rr{95.99} & \rr{96.42} & \rr{96.98} & \rr{80.55} & \rr{88.94} & \rr{90.96} & \rr{85.55} & \rr{86.29} & \rr{88.96} & \rr{87.59}\\
\cdashline{1-18}
Swin-UNETR$\dag$~\cite{tang2022self}     & 10k & 69.96 & 78.32 & 83.20 & 85.88 & 88.83 & 90.47 & 92.30 & 95.97 & 96.26 & 78.15 & 86.61 & 89.96 & 84.54 & 86.27 & 88.04 & 86.32\\
MAE3D$\dag$~\cite{chen2023masked}          & 10k & 70.59 & 79.44 & 83.52 & 85.78 & 89.13 & 90.75 & 95.57 & 96.19 & 96.92 & 77.01 & 86.94 & 88.55 & 84.44 & 86.27 & 88.50 & 86.64\\
\rowcolor{myred}
\textbf{MiM}         & 10k & \rr{75.34} & \rr{81.80} & \rr{85.61} & \rr{86.41} & \rr{89.77} & \rr{91.12} & \rr{96.36} & \rr{96.68} & \rr{96.99} & \rr{81.29} & \rr{89.06} & \rr{91.60} & \rr{86.29} & \rr{87.79} & \rr{89.67} & \rr{88.34}\\
\hline
\whline
    \end{tabular}
    }
    \vspace{-5pt}
    \label{tab:sota}
\end{table*}

\vspace{-10pt}
\subsection{Experiments on downstream tasks}

\subsubsection{Comparison on the BTCV dataset}
We first conducted experiments on the BTCV~\cite{landman2015miccai}, and the results are presented in Table~\ref{tab:btcv_benchmark}. Among the comparison methods, SimCLR~\cite{chen2020simple}, MoCov3~\cite{chenempirical}, DINO~\cite{caron2021emerging}, localMIM~\cite{wang2023masked}, HPM~\cite{wang2023masked}, MAE3D~\cite{chen2023masked}, GL-MAE~\cite{zhuang2023advancing} adopted UNETR~\cite{hatamizadeh2022unetr} as networks architecture. Most other methods, including our MiM, used Swin-UNETR~\cite{hatamizadeh2021swin}  as per the settings of previous works~\cite{tang2022self}. \ud{Table~\ref{tab:btcv_benchmark} includes the details of the backbone for 3D UNet~\cite{ronneberger2015unet}, UNETR~\cite{hatamizadeh2022unetr}, and Swin-UNETR~\cite{hatamizadeh2021swin}. UNETR~\cite{hatamizadeh2022unetr} uses ViT~\cite{dosovitskiy2020image} and Swin-UNETR~\cite{hatamizadeh2021swin} employs the Swin Transformer~\cite{liu2021swin}. For all experiments, we used ViT-Base~\cite{dosovitskiy2020image} for UNETR~\cite{hatamizadeh2022unetr} and Swin-Base~\cite{liu2021swin} for Swin-UNETR~\cite{hatamizadeh2021swin}, balancing performance and computational efficiency. These pre-trained encoders were used to initialize the encoder of respective segmentation networks.}

\textbf{Remark.} As shown in Table~\ref{tab:btcv_benchmark}, we observed that General SSL methods performed worse than Medical SSL methods. Specifically, SimCLR~\cite{chen2020simple} and MoCov3~\cite{chenempirical} achieved only 73.85\% and 79.54\%, respectively. This is because these methods rely on a large batch size and negative samples to avoid trivial constants, which is impractical for 3D medical images. Additionally, the negative relationship between different images used in SimCLR~\cite{chen2020simple} and MoCov3~\cite{chenempirical} is not suitable for 3D medical images. DINO~\cite{caron2021emerging} also achieved limited improvements. Our proposed MiM, outperformed MAE-based methods such as MAE3D~\cite{chen2023masked}, GL-MAE~\cite{zhuang2023advancing}, localMIM~\cite{wang2023masked}, HPM~\cite{wang2023masked}, and MCMAE~\cite{gao2022convmae} by a significant margin. We concluded that General SSL methods are not very suitable for 3D medical images, and it's crucial to consider the characteristics of 3D medical images when designing SSL methods.

The scratch Swin-UNETR~\cite{hatamizadeh2021swin} only achieves 80.53\% DSC. By pre-training MiM on a \ib{1k} unlabeled dataset, we gained a 3.93\% improvement with 84.46\% DSC, which outperforms existing methods by a clear margin. Among the compared methods, Swin-UNETR~\cite{tang2022self} and MAE3D~\cite{chen2023masked} achieved the best 82.58\% and second-best 82.40\% DSC, respectively. Our MiM surpasses these two methods by 1.88\% and 2.06\% DSC, respectively, which is a clear improvement on this dataset.

It's worth noting that scaling laws~\cite{kaplan2020scaling} also apply to 3D medical image pre-training. By pre-training with larger scale unlabeled \ib{10k} dataset, we observed that Swin-UNETR~\cite{tang2022self} and MAE3D~\cite{chen2023masked} achieved DSC scores of 83.20\% and 83.52\%, respectively. Our MiM with \ib{10k} achieved a DSC score of 85.41\%, which consistently outperformed these two methods significantly. These results suggest that scaling plays an important role in pre-training for 3D medical images and that our MiM method is effective for pre-training on larger datasets.

\noindent\textbf{Qualitative results.}
MiM was found to improve the completeness of segmentation results, as shown in Fig~\ref{fig:SegVisualizatoin}. The results of segmentation using MiM were better than existing methods.

\subsubsection{Comparison on the Unseen datasets}
We further conduct experiments on unseen datasets in pre-training, \textit{i.e.,} \ud{MM-WHS~\cite{zhuang2018multivariate}}, Spleen~\cite{simpson2019large}, Amos22~\cite{ji2022amos}, and Flare22~\cite{ma2023unleashing}. The results of these four datasets are shown in Table~\ref{tab:sota}. It can be observed that our MiM consistently outperformed all existing methods with a clear margin, which demonstrates promising generalizability to unseen datasets. Specifically, MiM outperformed existing methods by at least 1.89\% DSC on average. By pre-training with a larger scale of unlabeled dataset \ib{10k}, Swin-UNETR~\cite{tang2022self} and MAE3D~\cite{chen2023masked} improved by 0.92\% and 0.94\% DSC with 86.32\% and 86.64\%, respectively. Our MiM also gained improvement to 88.34\% DSC and outperformed these two methods consistently. \ud{MiM also shows} label efficiency when finetuning with less labels~\cite{tang2022self}. Specifically, MiM with 50\% label achieved comparable performance with the scratch Swin-UNETR with 100\% label with a clear margin.

\begin{table*}[!htb]
    \vspace{-5pt}
    \centering
    \caption{Experiment results on Five CT-based tasks on MSD dataset~\cite{simpson2019large}.}
      \vspace{-8pt}
    \resizebox{1\linewidth}{!}{
    \begin{tabular}{lcc|cc|cc|cc|cc|cc|cc}
        \whline
        \multirow{2}{*}{Method} & \multirow{2}{*}{\makecell[c]{Pre-training \\dataset}} & \multirow{2}{*}{Network} &\multicolumn{2}{c|}{Task03 Liver} & \multicolumn{2}{c|}{Task06 Lung} & \multicolumn{2}{c|}{Task07 Pancreas} & \multicolumn{2}{c|}{Task08 Hepatic Vessel} & \multicolumn{2}{c|}{Task10 Colon} & \multicolumn{2}{c}{\textbf{Avg}}\\
        \cline{4-15}
        ~ & & & DSC(\%) & NSD(\%) & DSC(\%) & NSD(\%) & DSC(\%) & NSD(\%)  & DSC(\%) & NSD(\%) & DSC(\%) & NSD(\%) & DSC(\%) & NSD(\%)\\
        \hline
        \textit{\textbf{From scratch}}\\
3D UNet~\cite{ronneberger2015unet}     & -  & -          & 94.41\s{1.45} & 93.94\s{2.64} & 48.09\s{9.41} & 42.65\s{6.89} & 75.33\s{3.06} & 90.98\s{3.50} & 56.11\s{2.29} & 77.58\s{2.24} & 40.97\s{3.08} & 54.63\s{4.37} & 62.98 & 71.95\\
UNETR~\cite{hatamizadeh2022unetr}       & -  & -          & 94.27\s{1.78} & 94.00\s{2.80} & 44.26\s{6.75} & 40.74\s{8.83} & 74.12\s{2.89} & 88.79\s{3.42} & 58.45\s{1.46} & 79.34\s{2.21} & 27.39\s{3.37} & 42.55\s{6.32} & 59.70 & 69.08\\
Swin-UNETR~\cite{hatamizadeh2021swin}  & -  & -          & 94.52\s{1.40} & 94.08\s{2.87} & 51.97\s{2.16} & 48.68\s{4.24} & 75.75\s{2.23} & 88.87\s{3.22} & 57.24\s{3.37} & 77.84\s{2.76} & 44.57\s{5.20} & 56.15\s{6.66} & 64.84 & 73.12\\
        \hline
        \textit{\textbf{General SSL}} \\
MoCov3~\cite{chenempirical}      & 1k & UNETR        & 94.15	   & 93.22       & 52.11	 & 48.16         & 74.16	 & 78.13         & 58.11	 & 78.13         & 43.99	 & 55.13         & 64.90 & 72.55 \\
localMIM~\cite{wang2023masked}    & 1k & UNETR        & 94.72	   & 93.69       & 52.93	 & 48.75         & 74.74	 & 79.77         & 58.86	 & 78.77         & 44.52	 & 54.35         & 65.55 & 73.10\\
HPM~\cite{wang2023masked}         & 1k & UNETR        & 94.63	   & 93.25       & 53.63	 & 48.34         & 74.74	 & 78.13         & 59.89	 & 78.93         & 42.67	 & 55.13         & 65.51 & 72.81\\
SimMIM~\cite{xie2022simmim}      & 1k & Swin-UNETR   & 94.50	   & 92.45       & 56.93	 & 52.41         & 74.99	 & 79.77         & 60.44	 & 78.99         & 42.42	 & 54.35         & 66.26 & 73.65\\
MCMAE~\cite{gao2022convmae}       & 1k & Swin-UNETR   & 95.49	   & 94.33       & 57.08	 & 57.18         & 75.78	 & 78.13         & 60.44	 & 78.93         & 45.63	 & 55.13         & 67.28 & 74.96\\
        \hline
        \textit{\textbf{3D Medical SSL}} \\
ModelGen~\cite{zhou2021models}    & 1k & 3D Unet    & 94.49\s{1.60} & 94.22\s{3.67} & 56.26\s{4.35} & 51.80\s{8.19} & 76.92\s{3.16} & 91.70\s{3.03} & 58.28\s{2.05} & 78.34\s{2.89} & 47.15\s{4.46} & 58.08\s{6.29} & 66.68 & 74.83\\
PCRLv2~\cite{zhou2023unified}      & 1k            & 3D Unet    & 94.61\s{1.43} & 95.78\s{1.95} & 62.57\s{6.58} & 59.67\s{8.52} & 77.02\s{2.64} & 92.67\s{3.11} & 53.63\s{1.75} & 76.25\s{2.60} & 39.78\s{8.13} & 53.50\s{9.98} & 65.52 & 75.57\\
GL-MAE~\cite{zhuang2023advancing}      & 1k & UNETR      & 94.65	   & 93.48         & 53.28	   & 49.51         & 72.96	   & 87.90         & 60.26	   & 78.65         & 45.52	   & 59.45         & 65.53 & 74.20\\
MAE3D~\cite{chen2023masked}       & 1k            & UNETR      & 95.53\s{0.92} & 96.62\s{1.44} & 58.30\s{12.0} & 52.32\s{13.0} & 77.31\s{2.96} & 91.92\s{3.50} & 60.44\s{2.23} & 78.80\s{2.69} & 45.40\s{0.05} & 56.00\s{5.83} & 66.76 & 75.33\\

Rubik++~\cite{zhu2020rubik}     & 1k & Swin-UNETR & 94.49\s{1.60} & 94.22\s{3.67} & 56.43\s{4.39} & 52.84\s{4.90} & 76.14\s{2.91} & 90.88\s{3.19} & 60.42\s{1.79} & 78.71\s{2.52} & 40.49\s{5.37} & 50.78\s{5.21} & 65.63 & 73.69\\
GVSL~\cite{he2023geometric}        & 1k & Swin-UNETR & 95.62\s{1.61} & 95.78\s{1.95} & 59.44\s{4.92} & 54.59\s{5.36} & 77.82\s{2.23} & 92.61\s{2.42} & 57.74\s{3.45} & 77.93\s{4.02} & 50.55\s{3.71} & 61.42\s{5.24} & 68.23 & 76.76\\
Swin-UNETR~\cite{tang2022self}  & 1k & Swin-UNETR & 95.55\s{0.33} & 96.61\s{0.78} & 60.23\s{1.86} & 57.22\s{0.19} & 76.93\s{2.68} & 90.88\s{3.36} & 60.44\s{2.32} & 79.08\s{2.22} & 43.21\s{5.38} & 55.43\s{6.90} & 68.98 & 75.16\\
\rowcolor{myred}
\textbf{MiM}& 1k & UNETR      & 96.01         &  97.03        & 63.06         & 60.11         & 78.39         & 92.71         & 60.71         & 79.44          & 51.02        & 62.57          & 69.79 & 78.37\\
\rowcolor{myred}
\textbf{MiM}& 1k & Swin-UNETR & \rr{96.14}\s{0.28} & \rr{97.43}\s{0.40} & \rr{63.52}\s{7.46} & \rr{60.92}\s{10.17} & \rr{78.86}\s{2.49} & \rr{92.80}\s{2.75} & \rr{60.85}\s{2.21} & \rr{79.55}\s{2.87} & \rr{51.02}\s{xx.xx} & \rr{63.03}\s{xx.xx} & \rr{70.07} & \rr{78.75}\\
        \cdashline{1-15}
        \rowcolor{myred}
\textbf{MiM}& 10k & Swin-UNETR & \rr{96.41\s{0.28}} & \rr{97.63\s{3.23}} & \rr{64.46\s{5.34}} & \rr{62.59\s{9.21}} & \rr{79.19\s{2.04}} & \rr{93.14\s{2.75}} & \rr{61.72\s{2.58}} & \rr{80.06\s{2.53}} & \rr{52.73}\s{3.53} & \rr{64.85\s{3.88}} & \rr{70.76} & \rr{79.67}\\
        \whline
        \end{tabular}
    }
	\vspace{-5pt}
    \label{tab:transferMSD}
\end{table*}

\subsubsection{Comparison on the MSD datasets}
To evaluate the generalizability on organ segmentation tasks, we conducted experiments on the five CT-based tasks on the MSD dataset~\cite{simpson2019large}, \textit{i.e.,} Task 03 Liver, Task06 Lung, Task07 Pancreas, Task08 Hepatic Vessel and Task10 Colon. Since existing methods didn't conduct experiments with the same pre-training dataset, we reimplemented these methods for fair comparison. It can be observed in Table~\ref{tab:transferMSD} that MiM achieves the best average DSC (70.07\%) and NSD(78.75\%) in all tasks. Since the scratch Swin-UNETR~\cite{hatamizadeh2021swin} performs better \ud{than} UNETR~\cite{hatamizadeh2022unetr} in terms of avg DSC (64.84\% vs 62.98\%) and NSD (73.12\% vs 69.08). We further pre-trained UNETR~\cite{hatamizadeh2022unetr} based on MiM for a fair comparison. It can be observed that with MiM pre-training, Swin-UNETR~\cite{hatamizadeh2021swin} gained an average improvement 5.23\% and 5.63\% in terms of DSC and NSD, respectively. With UNETR~\cite{hatamizadeh2022unetr} as the network, we observed an average improvement of 10.09\% and 9.29\% in terms of DSC and NSD, respectively. Moreover, by pre-training with a larger scale of unlabeled dataset \ib{10k}, MiM further improved to 70.76\% and 79.67\% in DSC and NSD, respectively.

\subsubsection{Comparison on CC-CCII dataset}
To evaluate the generalizability of our MiM on classification task, we fine-tuned it on the CC-CCII~\cite{zhang2020clinically} dataset and compared its performance with state-of-the-art general and medical SSL methods. Since existing SSL methods didn't conduct experiments on this dataset, we reproduced the related methods and reported the results. As shown in Table~\ref{tab:transferCCIICC}, it can be observed that MiM achieved the best performance in terms of ACC and AUC, surpassing all other methods with 93.63\% and 99.39\%. These findings demonstrate that the learned representation by MiM can be well transferable to classification problems and can be used effectively for medical image classification tasks. With a larger scale of pre-training dataset \ib{10k}, our MiM further improved to 94.12\% and 99.52\% in ACC and AUC, respectively, which shows the scalability of our proposed method when transferring across tasks.
\begin{table}[htb]
    \centering
    \caption{Experiment results on CC-CCII dataset~\cite{zhang2020clinically}.}
      \vspace{-8pt}
    \resizebox{0.84\linewidth}{!}{
    \begin{tabular}{lcccc}
        \whline
\multirow{2}{*}{Method} & \multirow{2}{*}{\makecell[c]{Pre-traning \\dataset}} & \multirow{2}{*}{Network} & \multicolumn{2}{c}{Disease classification}\\
\cline{4-5}
~                                       & & ~                        & ACC(\%) & AUC\\
\hline
\textit{\textbf{From Scratch}}\\
ResNet~\cite{he2016deep}       & -        & -       & 85.94\s{2.25} & 87.61\s{1.66}\\
ViT~\cite{dosovitskiy2020image}          & -        & -       & 85.92\s{0.92} & 96.26\s{0.90}\\
Swin-ViT~\cite{liu2021swin}     & -        & -       & 87.15\s{0.77} & 97.32\s{0.98}\\
\hline
\textit{\textbf{General SSL}} \\
SimCLR~\cite{chen2020simple}       & 1k       & ResNet   & 87.12         & 96.21\\
MoCov3~\cite{chenempirical}       & 1k       & ResNet   & 87.01         & 95.49\\
localMIM~\cite{wang2023masked}     & 1k       & ViT      & 88.15         & 97.58\\
HPM~\cite{wang2023hard}         & 1k       & ViT      & 88.26         & 97.65\\
SimMIM~\cite{xie2022simmim}       & 1k       & Swin-ViT & 89.62         & 98.16\\
MCMAE~\cite{gao2022convmae}        & 1k       & Swin-ViT & 90.26         & 97.12\\
\hline
\textit{\textbf{Medical SSL}} \\
PCRLv1~\cite{zhou2021preservational}       & 1k       & ResNet   & 88.84         & 97.69\\
PCRLv2~\cite{zhou2023unified}       & 1k       & ResNet   & 89.35         & 98.05\\
GL-MAE~\cite{zhuang2023advancing}       & 1k       & ViT      & 88.00\s{0.32} & 96.97\s{0.07}\\
MAE3D~\cite{chen2023masked}        & 1k       & ViT      & \bb{91.30\s{2.68}} & \bb{98.13\s{0.99}}\\
Rubik++~\cite{zhu2020rubik}      & 1k       & Swin-ViT & 89.93         & 98.55\\
SwinMM~\cite{wang2023swinmm}       & 1k       & Swin-ViT & 89.99         & 98.77\\
Swin-UNETR~\cite{tang2022self}   & 1k       & Swin-ViT & 91.81\s{0.26} & 98.97\s{0.08}\\
\rowcolor{myred}
\textbf{MiM} & 1k       & Swin-ViT & \rr{93.63\s{0.12}} & \rr{99.39\s{0.13}}\\
\cdashline{1-5}
\rowcolor{myred}
\textbf{MiM} & 10k      & Swin-ViT & \rr{94.26} & \rr{99.69}\\

        \whline
        \end{tabular}
    }
	\vspace{-5pt}
    \label{tab:transferCCIICC}
\end{table}

\subsubsection{Comparison on the BraTS 21 dataset}
To evaluate the generalizability of our MiM on MRI datasets, we fine-tuned it on the BraST 21~\cite{simpson2019large} MRI tumor segmentation dataset and compared its performance with state-of-the-art general and medical SSL methods. WT, TC, and ET denote the whole tumor, tumor core, and enhancing tumor, respectively.
It can be observed in Table~\ref{tab:transferMri} that both SSL methods can improve the performance of the model in segmenting tumors in BrasTS 21~\cite{simpson2019large} dataset. This is because CT and MRI are often used for the same task but for different purposes, and thus share similar anatomical structures. Therefore, the knowledge learned by SSL methods from unlabeled CT datasets can be transferred to MRI datasets~\cite{lyu2022conversion,tang2022self}. Our MiM outperformed all other methods by at least 1.34\% with 79.28\% DSC. By pre-training with a larger unlabeled dataset \ib{10k}, our MiM further improved to 79.92\%, which shows the scalability of our proposed method when transferring across modalities.
\begin{table}[htb]
    \centering
    \vspace{-5pt}
    \caption{Experiment results on BraTS-21~\cite{simpson2019large}. }\label{tab:transferMri}
    \vspace{-8pt}
    \resizebox{0.96\linewidth}{!}{
    \begin{tabular}{lcccccc}
    \hline
        \whline
        \multirow{2}{*}{Method} & \multirow{2}{*}{\makecell[c]{Pre-traning \\dataset}} & \multirow{2}{*}{Network} & \multicolumn{4}{c}{Dice Score(\%)} \\
        \cline{4-7}
                                &                       &   & TC    & WT    & ET    & \textbf{AVG}\\
        \hline
        \textit{\textbf{From Scratch}}\\
UNETR~\cite{hatamizadeh2022unetr}                   & -  & -                        & 81.62 & 87.81 & 57.34 & 75.58\\
Swin-UNETR~\cite{hatamizadeh2021swin}              & -  & -                        & 81.28 & 88.67 & 57.73 & 75.89\\
        \hline
        \textit{\textbf{General SSL}} \\
MoCov3~\cite{chenempirical}                  & 1k & UNETR                    & 82.60 & 88.89 & 57.69 & 76.39\\
localMIM$\dag$~\cite{wang2023masked}          & 1k & UNETR                    & 82.44 & 88.78 & 58.64 & 76.62\\
SimMIM$\dag$~\cite{xie2022simmim}            & 1k & Swin-UNETR               & 84.06 & 90.43 & 59.07 & 77.85\\
MCMAE$\dag$~\cite{gao2022convmae}             & 1k & Swin-UNETR               & 84.27 & 90.52 & 59.04 & 77.94\\
        \hline
        \textit{\textbf{3D Medical SSL}} \\
MAE3D$\dag$~\cite{chen2023masked}             & 1k & UNETR                    & 82.34 & \bb{90.35} & \bb{59.18} & \bb{77.29}\\
PCRLv2~\cite{zhou2023unified}                  & 1k & Swin-UNETR               & 82.13 & 90.06 & 57.70 & 76.63\\
Swin-UNETR~\cite{tang2022self}              & 1k & Swin-UNETR               & 82.51 & 89.08 & 58.15 & 76.58\\
SwinMM~\cite{wang2023swinmm}                  & 1k & Swin-UNETR               & 83.48 & 90.47 & 58.72 & 77.56\\
        \rowcolor{myred}
\textbf{MiM}            & 1k & Swin-UNETR               & \rr{84.99} & \rr{91.92} & \rr{60.94} & \rr{79.28}\\
\cdashline{1-5}
        \rowcolor{myred}
\textbf{MiM}            & 10k& Swin-UNETR               & \rr{85.96} & \rr{92.34} & \rr{61.45} & \rr{79.92}\\
        \whline
    \end{tabular}
    }
    \vspace{-5pt}
\end{table}

\subsection{Analysis of our proposed method}
All the models were pre-trained on \ib{1k}, which were then evaluated on BTCV~\cite{landman2015miccai} and MM-WHS~\cite{zhuang2018multivariate}.

\subsubsection{Ablation study}\label{sec:ablation}
\noindent\textbf{Loss functions.} \ud{We conducted comprehensive ablation studies on the BTCV~\cite{landman2015miccai} and MM-WHS~\cite{zhuang2018multivariate} validation datasets to evaluate the effectiveness of our hierarchical design, focusing on multi-level reconstruction and cross-level alignment components. As shown in Table~\ref{tab:ablation_study}, the multi-level reconstruction losses $L_R^1$, $L_R^2$, and $L_R^3$ substantially enhance model performance by capturing features at multiple granularities from 3D medical images, outperforming traditional single-level mask auto-encoding approaches (Rows 3-4). The importance of proper loss function composition is evident when examining the cross-level alignment loss $L_C$ in isolation (Row 2), which resulted in poor convergence with modest performance: 81.92\% DSC, 77.11\% NSD on BTCV and 89.99\% DSC, 74.12\% NSD on MM-WHS. This observation indicates that reconstruction losses $L_R$ play a fundamental role in establishing robust within-level representations before cross-level alignment $L_C$ can be effectively applied. Further analysis reveals that alignment $L_C^{2,3}$ between finer levels $x^2$ and $x^3$ (Row 5) shows modest gains since fine-grained details are primarily addressed by reconstruction losses, whereas alignment $L_C^{1,2}$ between coarser levels $x^1$ and $x^2$ (Row 6) yields more substantial improvements in both DSC and NSD metrics by effectively capturing global context. The optimal performance across both datasets was achieved by combining all reconstruction and cross-level alignment losses (Last row), demonstrating the synergistic relationship between multi-level reconstruction and cross-level alignment in hierarchical representation learning.}
\begin{table}[!ht]
    \centering
    \vspace{-5pt}
    \caption{Evaluation of the loss terms $\mathcal{L}_{R}$ and $\mathcal{L}_{C}$.}
    \vspace{-5pt}
    \resizebox{1\linewidth}{!}{%
    \begin{tabular}{ccccc|cc|cc}
            \whline
\multicolumn{5}{c|}{\textbf{Loss}} & \multicolumn{2}{c|}{BTCV} & \multicolumn{2}{c}{MM-WHS}\\
            \hline
$\ud{\mathcal{L}_\mathcal{R}^1}$ & $\ud{\mathcal{L}_\mathcal{R}^2}$ & $\ud{\mathcal{L}_\mathcal{R}^{3}}$ &$\ud{\mathcal{L}_\mathcal{C}^{12}}$ & $\ud{\mathcal{L}_\mathcal{C}^{23}}$ & DSC(\%) & NSD(\%) & DSC & NSD(\%)\\
            \hline
            \rowcolor{mygray}
\XSolidBrush & \XSolidBrush & \XSolidBrush & \XSolidBrush & \XSolidBrush& 81.49\s{0.43} & 75.92\s{0.89}                                  & 89.42\s{0.77}                                          &  73.06\s{0.18}                                     \\
\ud{\XSolidBrush}   & \ud{\XSolidBrush}   & \ud{\XSolidBrush} & \ud{\CheckmarkBold} & \ud{\CheckmarkBold}     &
\ud{\tabincell{c}{81.92\greenp{0.43$\uparrow$}}} & \ud{\tabincell{c}{77.11\greenp{1.91$\uparrow$}}} & \ud{\tabincell{c}{89.99\greenp{0.57$\uparrow$}}} & \ud{\tabincell{c}{74.12\greenp{1.06$\uparrow$}}} \\
\CheckmarkBold   & \XSolidBrush & \XSolidBrush & \XSolidBrush & \XSolidBrush &
\tabincell{c}{83.10\s{0.09}\greenp{1.60$\uparrow$}} & \tabincell{c}{79.51\s{0.04}\greenp{3.59$\uparrow$}} & \tabincell{c}{90.44\s{0.20}\greenp{1.02$\uparrow$}} & \tabincell{c}{74.53\s{0.45}\greenp{1.47$\uparrow$}} \\
\CheckmarkBold   & \CheckmarkBold   & \CheckmarkBold   & \XSolidBrush & \XSolidBrush &
\tabincell{c}{84.25\s{0.21}\greenp{2.76$\uparrow$}} & \tabincell{c}{81.29\s{5.37}\greenp{5.37$\uparrow$}} & \tabincell{c}{90.74\s{0.07}\greenp{1.32$\uparrow$}}     & \tabincell{c}{75.50\s{0.03}\greenp{2.44$\uparrow$}}\\
\cdashline{1-9}
\CheckmarkBold   & \CheckmarkBold   & \CheckmarkBold & \XSolidBrush & \CheckmarkBold     &
\tabincell{c}{84.29\s{0.67}\greenp{2.80$\uparrow$}} & \tabincell{c}{81.45\s{0.63}\greenp{5.53$\uparrow$}} & \tabincell{c}{90.89\s{0.05}\greenp{1.47$\uparrow$}} & \tabincell{c}{75.97\s{0.51}\greenp{2.91$\uparrow$}} \\
\CheckmarkBold   & \CheckmarkBold   & \CheckmarkBold & \CheckmarkBold & \XSolidBrush     &
\tabincell{c}{84.43\s{0.52}\greenp{2.93$\uparrow$}} & \tabincell{c}{81.56\s{0.23}\greenp{5.63$\uparrow$}} & \tabincell{c}{90.89\s{0.01}\greenp{1.47$\uparrow$}} & \tabincell{c}{76.27\s{0.36}\greenp{3.20$\uparrow$}} \\
\rowcolor{myred} \CheckmarkBold & \CheckmarkBold   & \CheckmarkBold   & \CheckmarkBold   & \CheckmarkBold   &
\tabincell{c}{84.66\s{0.35}\greenp{3.16$\uparrow$}} & \tabincell{c}{82.11\s{0.30}\greenp{6.19$\uparrow$}} & \tabincell{c}{91.04\s{0.10}\greenp{1.62$\uparrow$}} & \tabincell{c}{76.65\s{0.43}\greenp{3.59$\uparrow$}}\\
            \whline
        \end{tabular}
        }
    \label{tab:ablation_study}
\end{table}

\noindent\ud{\textbf{Patch Types for Generating subsequent level volumes.}\label{sec:recontruction_target} During the patchification process (Figure~\ref{fig:mim}), each level-$l$ volume ($x^l$) is divided into masked and unmasked patches. We evaluate both patch types for generating the subsequent level-$l+1$ volume ($x^{l+1}$) in our framework. As shown in Table~\ref{tab:ablation_reconstruction_target}, using masked patches from $x^{l}$  to generate $x^{l+1}$ consistently achieves better performance than using unmasked patches from $x^{l}$. This superiority stems from masked patches forcing the model to recover missing information, thereby promoting effective reconstruction of overlapping regions across levels and learning of multi-scale semantic representations. In contrast, unmasked patches directly expose original volume information to the model during the same iteration, effectively creating an information shortcut. This shortcut reduces the learning challenge by allowing the model to simply copy unmasked features rather than learning to infer and reconstruct them, ultimately limiting its ability to capture rich semantic details and develop robust generalization capabilities.}
\begin{table}[ht]
    \centering
    \vspace{-5pt}
    \caption{Evaluation of the \ud{patch types derived from level-l volumes $x^{l}$s utilized for subsequent level volumes $x^{l+1}$s. 
    We report the DSC(\%) and NSD(\%) on BTCV~\cite{landman2015miccai} and MM-WHS~\cite{zhuang2018multivariate}.}}
    \vspace{-5pt}
    \resizebox{1\linewidth}{!}{%
        \begin{tabular}{c|cc|cc}
            \whline
            \multirow{2}{*}{\ud{Patch Types for Generating $x^{l+1}$}} 
            & \multicolumn{2}{c|}{BTCV} 
            & \multicolumn{2}{c}{MM-WHS} \\
            \cline{2-5}
            & DSC (\%)          & NSD (\%) 
            & DSC (\%)          & NSD (\%) \\
            \hline
            Unmasked patches \ud{from $x^l$} 
            & 82.18             & 76.22 
            & 89.33             & 73.65 \\
            \rowcolor{myred}
            Masked patches \ud{from $x^l$}   
            & 84.66\s{0.23}     & 82.11 
            & 91.04\s{0.20}     & 76.65 \\
            \whline
        \end{tabular}
    }
    
    \label{tab:ablation_reconstruction_target}
\end{table}

\noindent\textbf{Negative pairs for $\mathcal{L}_{\mathcal{C}}$.}
In Eq.~\ref{eq:cross_loss}, we employ infoNCE loss\cite{chen2020simple} as the default choice. This loss maximizes the similarity between the cross-level images and pushes away the negative samples. Another loss function is BYOL cosine loss~\cite{grill2020bootstrap}. The primary distinction is whether to use negative samples. As in Table~\ref{tab:ablation_cross}, negative samples aid in learning better representations~\cite{chen2020simple}, and thus, we include them in our default choices.
\begin{table}[ht]
    \centering
    \vspace{-5pt}
    \caption{Evaluation of $\mathcal{L}_{C}$ whether using negative samples.}
    \vspace{-8pt}
    \resizebox{0.85\linewidth}{!}{%
    \begin{tabular}{l|cc|cc}
        \whline
\multirow{2}{*}{Loss function $L_C$} & \multicolumn{2}{c|}{BTCV} & \multicolumn{2}{c}{MM-WHS} \\
\cline{2-5}
& DSC(\%)           & NSD(\%) & DSC(\%)           & NSD(\%)\\
\hline
BYOL-style cosine loss~\cite{grill2020bootstrap}     & 83.68\s{0.06} & 80.12 & 90.53\s{0.14} & 74.66\\
\rowcolor{myred}
Contrastive loss~\cite{chen2020simple}           & 84.66\s{0.23} & 82.11 & 91.04\s{0.20} & 76.65\\
\whline
    \end{tabular}}
    \label{tab:ablation_cross}
    \vspace{-5pt}
\end{table}

\noindent\textbf{Efficiency of MiM.}\label{sec:ablation_backbone}
\ud{This study adopts a hybrid convolution-transformer backbone that integrates convolutional blocks with transformer layers. By incorporating convolutional blocks, the architecture enhances inductive bias learning and enables the reuse of multi-scale features, effectively supporting hybrid representation learning~\cite{gao2022convmae}. This design enhances the transformer’s capability to process 3D medical images with greater efficiency and effectiveness. As such, this backbone is a foundational component of our proposed framework.} Table~\ref{tab:ablation_backbone} presents a comparative analysis of computation costs \textit{i.e.} flops and times during the pre-training period across various methodologies. Our evaluation highlights the computational efficiency of MAE-based approaches when coupled with UNETR, as they leverage only the \ud{unmasked} tokens from masked 3D medical images. In contrast, alternative MAE-based methods naively cooperating with hybrid backbones such as Swin-UNETR, necessitate processing all tokens. Our proposed method extends the hybrid backbone architecture~\cite{gao2022convmae} to 3D medical images in the pre-training stage, achieving significantly superior performance while maintaining computational efficiency.

\begin{table}[th]
    \begin{center}
    \vspace{-5pt}
    \caption{Evaluation of model complexity during pre-training and subsequent performance on downstream tasks. }
    \label{tab:ablation_backbone}
    \resizebox{1\linewidth}{!}{
    \begin{tabular}{ l|c|cc||cc||cc}
    \whline
    \multirow{2}{*}{Method} & \multirow{2}{*}{Networks}   & \multicolumn{2}{c||}{Tokens} & \multirow{2}{*}{flops(G)} &  \multirow{2}{*}{\tabincell{c}{Pre-training\\ duration\ud{(h)}}} & \multicolumn{2}{c}{\ud{DSC(\%)}}\\
    \cline{3-4} \cline{7-8}
    & & All & Partial & & & BTCV & MM-WHS\\
    \hline
    MAE3D             & UNETR      &  & \CheckmarkBold  & 12.3     & 9         & 82.20 & 90.03\\
    HPM 	          & UNETR      &  & \CheckmarkBold  & 34.1 	   & 9.9       & 82.04 & 89.86\\
    localMIM	      & UNETR      &  & \CheckmarkBold  & 26.6     & 10.3      & 81.38 & 89.99\\
    \hline
    Adam  		      & Swin-UNETR & \CheckmarkBold  &  & 43.7    & 27.0       & 80.45 & 88.82\\
    SimMIM     	      & Swin-UNETR & \CheckmarkBold  &  & 70.8    & 26.6       & \bb{82.24} & 90.22\\
    Swin-UNETR      & Swin-UNETR & \CheckmarkBold  &  & 394.84  & 130.0      & 82.58 & 90.32\\
    \rowcolor{myred}
    \textbf{MiM} 	  & Swin-UNETR & &  \CheckmarkBold & 45.8   & 14.6        & \rr{84.66} & \rr{91.04}\\
    \hline
    \whline
    \end{tabular}}
    \end{center}
\end{table}

\subsubsection{Hyper-parameter analysis}\label{sec:hyperparameter}

\begin{figure}[htbp]
\centering
\begin{minipage}[t]{0.46\linewidth}
\centering
\includegraphics[width=1\linewidth]{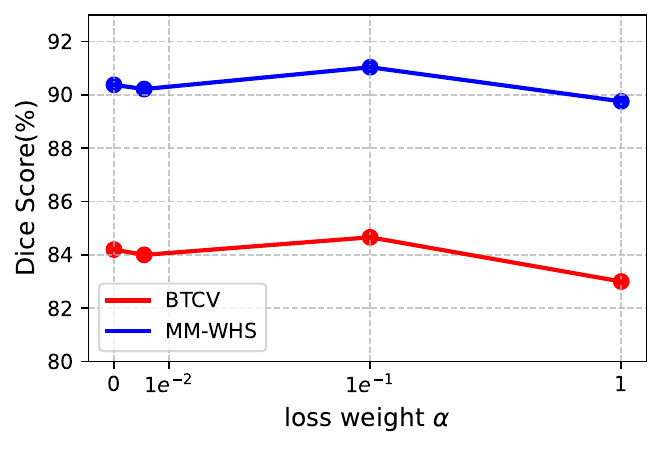}
	\vspace{-25pt}
\caption{\ud{Evaluation of the loss weight $\alpha$ for $\mathcal{L_{C}}$.}}
\label{fig:loss_weight}
\end{minipage}
\begin{minipage}[t]{0.46\linewidth}
\centering
\includegraphics[width=1\linewidth]{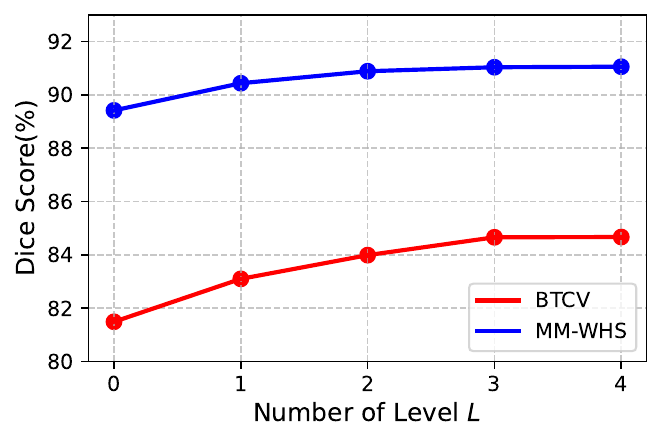}
	\vspace{-25pt}
\caption{Evaluation of the number of level $L$.}
	\label{fig:num_level}
\end{minipage}
\vspace{-10pt}
\end{figure}

\noindent\textbf{Weights of $\mathcal{L}_{\mathcal{R}}$ and $\mathcal{L}_{\mathcal{C}}$.}
In Eq.~\ref{eq:loss}, the loss function of MiM consists of two parts. Thus, in Fig~\ref{fig:loss_weight},
we increased the value of $\alpha$ . The optimal value of $\alpha$ was $1e^{-1}$ by observation. Further scaling up the weight of cross-level alignment did not bring any additional benefits. This may be because the magnitude of the cross-level alignment function is much larger than the magnitude of the reconstruction loss, resulting the ignorance of reconstruction process.

\noindent\textbf{Weight of different levels of $\mathcal{L}_{\mathcal{R}}$.}
\ud{To determine the optimal strategy for multi-level reconstruction, we evaluated four distinct learning processes during pretraining by varying the weights of different level reconstruction losses, as shown in Table~\ref{tab:ablation_masks}. Using the BTCV~\cite{landman2015miccai} and MM-WHS~\cite{zhuang2018multivariate} datasets, we found that our baseline model without hierarchical reconstruction - w/o Coarse \& Fine with all weights set to 0 - yielded the lowest performance with DSC scores of 81.4\% and 89.42\%,. The Coarse$\rightarrow$Fine process, which initially focused on coarser levels ($L_R^1$ and $L_R^2$) before prioritizing finer levels ($L_R^3$), improved performance to 83.99\% and 90.70\% DSC. Notably, the Fine$\rightarrow$Coarse process, which reversed this order by beginning with finer-level reconstruction, achieved even better results with DSC scores of 84.26\% and 90.85\%. The Simultaneously process, which maintained equal weights across all levels throughout training, emerged as the most effective approach, achieving the highest DSC scores of 84.66\% and 91.04\%. These results demonstrate that hierarchical learning strategies significantly enhance representation learning, with simultaneous multi-level reconstruction proving most beneficial.} 
\begin{table}[ht]
    \centering
    \vspace{-5pt}
    \caption{Evaluation of model with different learning process.}
    \vspace{-5pt}
    \resizebox{0.9\linewidth}{!}{%
    \begin{tabular}{l|ccc|cc}
        \whline
\multirow{2}{*}{Learning process}           &  \multicolumn{3}{c|}{\ud{Weight of Level-l}}            & \multicolumn{2}{c}{Datasets} \\
\cline{2-6}
                                            &  \ud{$\mathcal{L}_{R}^{1}$}   & \ud{$\mathcal{L}_{R}^{2}$}            & \ud{$\mathcal{L}_{R}^{3}$}           & BTCV           & MM-WHS \\
\hline
\rowcolor{mygray}
w/o Coarse \& Fine          & \multirow{4}{*}{1$\longrightarrow$1} & 0$\longrightarrow$0 & 0$\longrightarrow$0 & 81.49\s{0.43}  & 89.42\\
Coarse$\longrightarrow$Fine &                                      & 2$\longrightarrow$1 & 0$\longrightarrow$1 & 83.99\s{0.49}  & 90.70\\
Fine$\longrightarrow$Coarse &                                      & 0$\longrightarrow$1 & 2$\longrightarrow$1 & 84.26\s{0.73}  & 90.85\\
\rowcolor{myred}
Simultaneously                  &                                  & 1$\longrightarrow$1 & 1$\longrightarrow$1 & 84.66\s{0.35}  & 91.04\\
        \whline
    \end{tabular}}
	\vspace{-5pt}
    \label{tab:ablation_masks}
\end{table}

\begin{figure}[!h]
	\centering
	\includegraphics[width=1\linewidth]{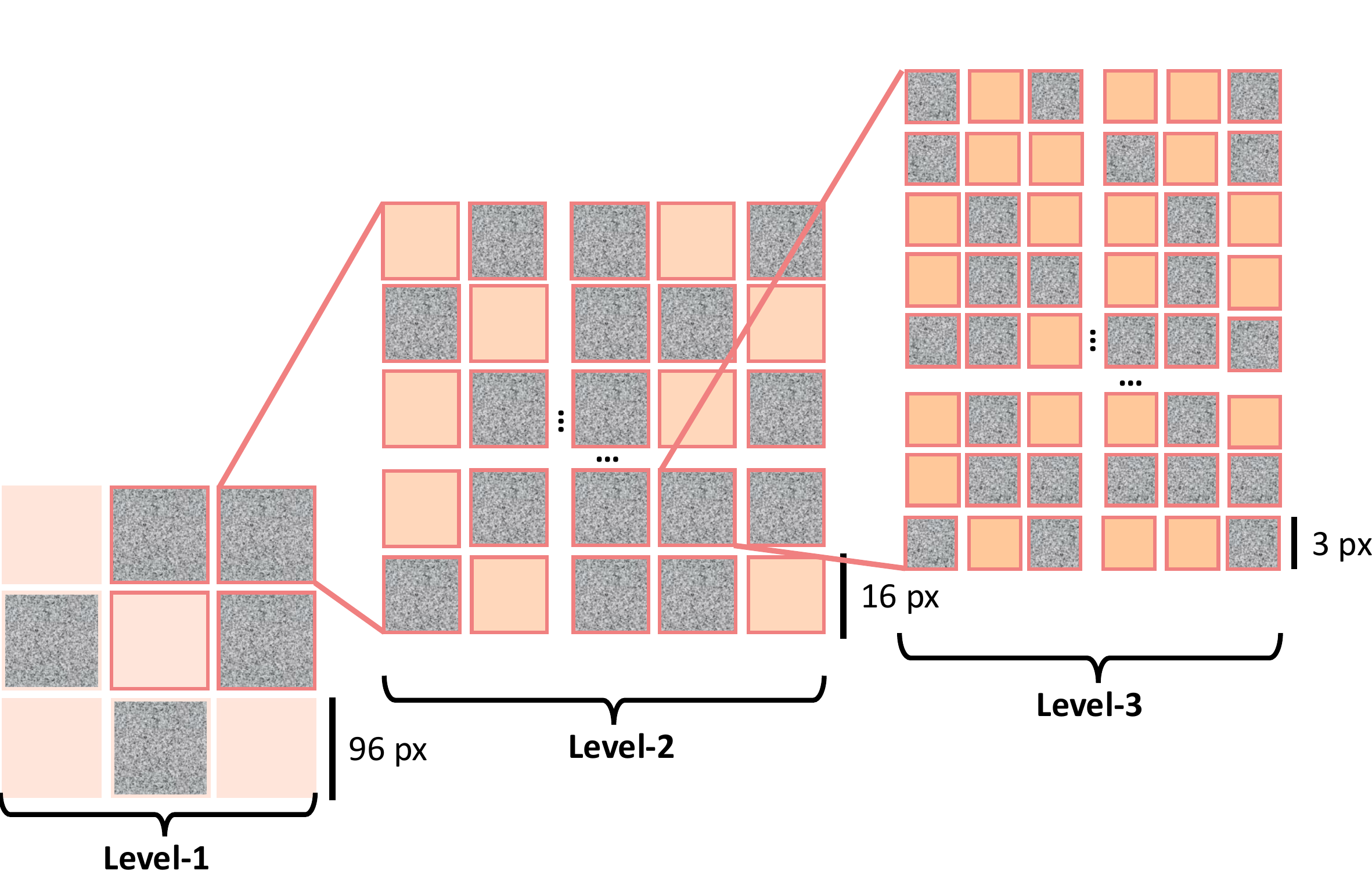}
	\vspace{-20pt}
	\caption{\ud{Illustration of our multi-level volume generation strategy. The framework processes volumes at three hierarchical levels with decreasing patch sizes: Level-1 (96 px patches), Level-2 (16 px patches), and Level-3 (3x3 patches). This design captures anatomical information at complementary scales while maintaining computational efficiency. Further subdivision beyond Level-3 would result in patches too small to preserve meaningful anatomical features.  Gray patches indicate masked regions, while color intensity (light to dark) represents the progression from coarse to fine granularity across three hierarchical levels.}}
	\label{fig:num_level_design}
\end{figure}

\noindent\textbf{Numbers of multi-levels $L$.} \ud{Our adoption of a three-level architecture is supported by both empirical evidence and practical considerations. Empirically, as demonstrated in Fig~\ref{fig:num_level_design}, our multi-level strategy preserves anatomical structures through carefully selected resolutions: Level-1 (384$\times$384$\times$192) captures broad contextual information, Level-2 (96$\times$96$\times$96) extracts intermediate features, and Level-3 (16$\times$16$\times$16) retains fine details. Through extensive experimentation shown in Fig~\ref{fig:num_level}, we investigated the impact of varying the number of levels $L$ and found that $L=3$ yields optimal performance. Adding a fourth level would result in an extremely low resolution ($1\times1\times1$), which, while potentially simplifying reconstruction, leads to suboptimal performance due to excessive information loss, consistent with findings in similar architectures\cite{he2022masked,xie2022simmim}.}

\subsubsection{Reconstruction results}
\begin{figure}[htbp]
    \centering
    \vspace{-5pt}
	  \includegraphics[width=0.6\linewidth]{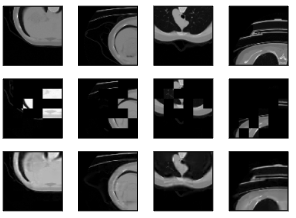}
    \vspace{-5pt}
    \caption{Reconstruction results on the  BTCV validation dataset~\cite{landman2015miccai}.}
    \label{fig:reconstruction}
    \vspace{-10pt}
\end{figure}

We provided reconstruction results of MiM on BTCV~\cite{landman2015miccai} in Fig.~\ref{fig:reconstruction}, where the first and second rows represent the volumes and masked 3D medical images, respectively, and the reconstruction results in the last row demonstrate excellent performance.

\subsubsection{\ud{Incorporating Parameter-Efficient Fine-Tuning Methods}}
\ud{Parameter-efficient fine-tuning methods, such as LoRA~\cite{hu2021lora} and Ladder fine-tuning~\cite{sung2022lst}, offer valuable insights in realistic scenarios with limited computation resources~\cite{hu2021lora,sung2022lst}. Specifically, we evaluated LoRA~\cite{hu2021lora} for both classification and segmentation tasks on 3D medical images using our proposed method, implementing it via the official codebase~\footnote{\url{https://github.com/microsoft/LoRA}}. Our implementation introduces low-rank matrices into attention and convolutional layers to approximate weight updates while freezing the original parameters during fine-tuning.
As demonstrated in Table~\ref{tab:tuning_methods_seg} and Table~\ref{tab:tuning_methods_cls}, LoRA achieves comparable performance to full fine-tuning while significantly reducing the number of updated parameters, consistent with previous findings~\cite{hu2021lora} and~\cite{sung2022lst}. We observed that increasing the rank $r$ improves performance, approaching that of full fine-tuning, though at the cost of increased training time and memory usage. These results show that our method can effectively leverage LoRA~\cite{hu2021lora} during fine-tuning, achieving an optimal balance between computational resources and performance.}
\begin{table}[h]
\centering
\caption{\ud{Comparison of different LoRA fine-tuning strategies applied to our method on the MM-WHS dataset~\cite{zhuang2018multivariate}. The Update Param. column shows the ratio between LoRA's trainable parameters and the total trainable parameters, excluding the segmentation decoder which was added later for downstream tasks.}}
\label{tab:tuning_methods_seg}
\vspace{-5pt}
\resizebox{1\linewidth}{!}{
\begin{tabular}{c|c|cc|cc}
\arrayrulecolor{mypink}\whline
\multirow{2}{*}{\ud{\makecell[c]{Fine-tuning strategies}}}             & \multirow{2}{*}{\ud{\makecell[c]{Update \\Param. (\%)}}} & \multicolumn{2}{c|}{\ud{Memory Usage (GB)}} & \multicolumn{2}{c}{\ud{Segmentation}} \\ 
\cline{3-6}
&                            & \ud{Train}  & \ud{Inference} & \ud{DSC(\%)}   & \ud{NSD(\%)}   \\ \whline
    \rowcolor{myred}
\ud{Fully fine-tuning(default)}            & \ud{100} & \ud{16.74} & \ud{0.81} & \ud{91.12} & \ud{96.19} \\ \hline 
\ud{LoRA~\cite{hu2021lora}(r=16)} & \ud{10.51} & \ud{17.03} & \ud{0.82} & \ud{91.09}  & \ud{96.72}\\ 
\ud{LoRA~\cite{hu2021lora}(r=8)}  & \ud{5.55} & \ud{16.72} & \ud{0.81} & \ud{90.97} &  \ud{96.58} \\ 
\ud{LoRA~\cite{hu2021lora}(r=4)}  & \ud{2.85} & \ud{16.58} & \ud{0.81} & \ud{91.01} & \ud{96.78}\\ 
\whline
\end{tabular}}

\end{table}

\begin{table}[h]
\centering
\caption{\ud{Comparison of different LoRA fine-tuning strategies applied to our method on the CC-CCII dataset~\cite{zhang2020clinically}. The Update Param. column shows the ratio between LoRA's trainable parameters and the total trainable parameters, excluding the classifier which was added later for classification downstream tasks.}}\label{tab:tuning_methods_cls}
\vspace{-5pt}
\resizebox{1\linewidth}{!}{
\begin{tabular}{c|c|cc|cc}
\arrayrulecolor{mypink}\whline
\multirow{2}{*}{\ud{\makecell[c]{Fine-tuning strategies}}} & \multirow{2}{*}{\ud{\makecell[c]{Update \\Param. (\%)}}} & \multicolumn{2}{c|}{\ud{Memory Usage (GB)}} & \multicolumn{2}{c}{\ud{Classification}} \\ 
\cline{3-6}
&                            & \ud{Train}  & \ud{Inference} & \ud{ACC(\%)}   & \ud{AUC}   \\ \whline
\rowcolor{myred}\ud{Fully fine-tuning(default)}   & \ud{100} & \ud{14.52} & \ud{0.20} & \ud{94.26} & \ud{99.69} \\ \hline
\ud{LoRA~\cite{hu2021lora}(r=16)} & \ud{11.17} & \ud{17.18} & \ud{0.21}                 & \ud{94.12}              & \ud{98.99}              \\ 
\ud{LoRA~\cite{hu2021lora}(r=8)}  & \ud{8.10}  & \ud{16.46} & \ud{0.21}                   & \ud{93.78}              & \ud{98.74}              \\ 
\ud{LoRA~\cite{hu2021lora}(r=4)}  & \ud{5.37}  & \ud{16.05} & \ud{0.21} & \ud{93.95} & \ud{98.78} \\ 
\whline
\end{tabular}}

\end{table}

\section{Conclusion}\label{sec:conclusion}

In this paper, we introduce the Mask in Mask (MiM) pre-training framework, which significantly advances 3D medical image analysis. 
\ud{By incorporating hierarchical designs, \textit{i.e.,} multi-level reconstruction and cross-level alignment, MiM efficiently encodes multi-granularity visual cues of structure and details into the representation.} To facilitate the fair and comprehensive comparison of existing methods, we collected ten public datasets and curated two scales of pre-training datasets, \textit{i.e.,} \ib{1k} and \ib{10k}. 
The results reveal that the hierarchical design of the MiM framework is crucial for achieving superior performance for 3D medical images. 
We further explore to scale up the pre-training dataset to \ib{10k}. The results show that the performance of MiM can be further improved by scaling up the pre-training dataset. This finding emphasizes the importance of large-scale pre-training for building the foundation model in 3D medical images. 

Several promising directions can be explored upon this work. (1) Building up a large-scale pre-training datasets, \textit{e.g.,} more than 100k volumes. (2) Exploring more downstream tasks, \textit{e.g.,} 3D medical image registration. \ud{(4) Exploring the potential of learnable level embeddings instead of hard coding the number of level.}
(3) Exploring cooperation with other modalities for multi-modal pre-training, \textit{e.g.,} language.
Actually, (1) is the cornerstone for the foundation model in 3D medical images, and (2) can further comprehend the evaluation. (3) can complement the 3D medical foundation model with the information from different modalities. (4) can further improve the method's design.


\bibliographystyle{IEEEtran}
\bibliography{bibs/abbr, bibs/pretrainingDataset,bibs/supervisedPretraining,bibs/ssl, bibs/model,bibs/datasets,bibs/optimization,bibs/generation}

\end{document}